\documentclass{article}
\usepackage{graphicx}
\usepackage{amsmath}
\usepackage{amssymb}
\usepackage{hyperref}
\usepackage{placeins}
\usepackage[edges]{forest}
\usepackage{geometry}
\usepackage{tikz}
\usepackage{pdflscape}
\usepackage{float}
\usepackage{booktabs}
\usetikzlibrary{shapes, arrows.meta, positioning}
\usepackage{graphicx}
\usepackage{afterpage}   
\usepackage{adjustbox}   

\title{AI Models for Depressive Disorder Detection and Diagnosis: A Review}

\author{%
	Dorsa Macky Aleagha\textsuperscript{1}\thanks{Email: \href{mailto:dorsa.macky@gmail.com}{dorsa.macky@gmail.com}},  
	Payam Zohari\textsuperscript{1}\thanks{Email: \href{mailto:payam.zohari@aut.ac.ir}{payam.zohari@aut.ac.ir}},  
	Mostafa Haghir Chehreghani\textsuperscript{1}\thanks{Corresponding author. Email: \href{mailto:mostafa.chehreghani@aut.ac.ir}{mostafa.chehreghani@aut.ac.ir}}
}

\date{\footnotesize\textsuperscript{\textbf{1}}Department of Computer Engineering,
	Amirkabir University
	of Technology (Tehran Polytechnic)\\
	Tehran, Iran}

\date{}	
	
\begin{document}

\maketitle


\begin{abstract}
		Major Depressive Disorder is one of the leading causes of disability worldwide, yet its diagnosis still depends largely on subjective clinical assessments. Integrating Artificial Intelligence (AI) holds promise for developing objective, scalable, and timely diagnostic tools. In this paper, we present a comprehensive survey of state-of-the-art AI methods for depression detection and diagnosis, based on a systematic review of 55 key studies. We introduce a novel hierarchical taxonomy that structures the field by primary clinical task (diagnosis vs. prediction), data modality (text, speech, neuroimaging, multimodal), and computational model class (e.g., graph neural networks, large language models, hybrid approaches). Our in-depth analysis reveals three major trends: the predominance of graph neural networks for modeling brain connectivity, the rise of large language models for linguistic and conversational data, and an emerging focus on multimodal fusion, explainability, and algorithmic fairness. Alongside methodological insights, we provide an overview of prominent public datasets and standard evaluation metrics as a practical guide for researchers. By synthesizing current advances and highlighting open challenges, this survey offers a comprehensive roadmap for future innovation in computational psychiatry.
\end{abstract}

\paragraph{keywords}
Depressive disorder, Artificial Intelligence (AI), multimodal analysis, Graph Neural Networks (GNNs), Large Language Models (LLMs), computational psychiatry


\section{Introduction}
Major Depressive Disorder (MDD), commonly known as depression, is a significant global health challenge characterized by a persistent low mood, a loss of interest in previously enjoyable activities (anhedonia), and a range of emotional, cognitive, and physical symptoms \cite{american2022dsm5}. As a leading cause of disability worldwide, depression contributes substantially to the overall global burden of disease, impacting not only individual well-being but also imposing considerable economic strain through reduced productivity and increased healthcare costs \cite{malhi2018depression, evans2021economic}. The early and accurate detection of depressive disorders is therefore crucial for initiating timely interventions and improving patient outcomes.

In recent years, the field of computational psychiatry has seen a surge of interest in leveraging Artificial Intelligence (AI) to address this challenge, moving beyond traditional, subjective diagnostic methods. The growing availability of large-scale digital datasets—spanning social media text, clinical interview transcripts, wearable sensor data, and neuroimaging—has created a fertile ground for the application of advanced machine learning models \cite{lethus2023artificial}. Despite this promise, significant challenges remain: integrating heterogeneous data modalities, ensuring model interpretability for clinical trust, and mitigating demographic biases that can lead to unfair predictions \cite{deng2024artificial, gichoya2022ai}. Nevertheless, sophisticated architectures such as Graph Neural Networks (GNNs) and Large Language Models (LLMs) are increasingly being deployed to capture complex patterns in brain connectivity and natural language, respectively.

Several valuable surveys have reviewed the intersection of AI and depression detection. However, they are often limited in scope, focus on a single modality or model family, and lack a structured framework for comparative analysis. Table~\ref{tab:related_surveys} summarizes the most relevant recent surveys and their limitations.

\begin{table*}[!t]
	\centering
	\caption{Summary of recent surveys on AI for depression detection.}
	\label{tab:related_surveys}
	\resizebox{\textwidth}{!}{
		\begin{tabular}{|l|l|l|p{7cm}|}
			\hline
			\textbf{Survey} & \textbf{Year} & \textbf{Focus} & \textbf{Limitations / Gap} \\ \hline
			Squires et.al. \cite{squires2023deep} & 2023 & ML/DL in psychiatry & Lacks a structured taxonomy; limited coverage of GNNs/LLMs. \\ \hline
			Subuhi and Vadlamudi \cite{subuhi2024advancements} & 2024 & ML/DL for depression & No hierarchical organization; no coverage of modern architectures. \\ \hline
			Ujalambkar \cite{ujalambkar2023comprehensive} & 2023 & Multimodal detection & Focuses on a single modality; proposes a model but lacks structural review. \\ \hline
			Kumar and Tyagi \cite{anonymous2023depression} & 2023 & Social network data & Reviews general ML/DL techniques without modality-specific organization. \\ \hline
			Jin et.al. \cite{jin2025applications} & 2025 & LLMs in mental health & Restricted to LLMs; does not integrate other model classes or modalities. \\ \hline
			Park et.al. \cite{ayano2024effectiveness} & 2024 & AI effectiveness & Focuses on AI's clinical effectiveness, not on methodological taxonomy. \\ \hline
			Yasin et.al. \cite{abdalrazaq2023machine} & 2023 & Audiovisual/EEG modalities & Modality-specific; lacks a cross-cutting hierarchical structure. \\ \hline
			Ren et.al. \cite{ren2025aiapplications} & 2025 & Bibliometric analysis & Purely bibliometric; no structured, qualitative synthesis of models. \\ \hline
		\end{tabular}
	}
\end{table*}

As the table makes clear, no existing survey offers a comprehensive, hierarchical taxonomy that systematically organizes the field first by the \textbf{clinical task} (diagnosis vs.\ prediction), then by \textbf{data modality} (text, speech, neuroimaging, multimodal), and finally by the \textbf{class of computational model} employed (e.g., GNNs, LLMs, hybrid approaches). This structural gap makes it difficult for researchers to navigate the landscape, compare methodologies across studies, and identify specific research opportunities. Moreover, prior surveys lack dedicated synthesis paragraphs that extract methodological patterns, contradictions, and consensus, and they provide no consolidated guide to evaluation metrics or public datasets.

\subsection{Our Contributions}
This survey aims to fill these critical gaps. Based on a systematic review of 55 key studies published between 2021 and 2025, our paper makes the following contributions:

\begin{enumerate}
	\item \textbf{A novel hierarchical taxonomy:} We introduce a structured, three-level framework that categorizes the literature by clinical task, data modality, and AI model class, enabling researchers to quickly locate relevant work and understand the field's structure at a glance.
	
	\item \textbf{Comprehensive and structured review:} We provide an in-depth analysis of state-of-the-art methods, following our taxonomy to discuss innovations and findings within each sub-domain. We include synthesis paragraphs at the end of each subsection that explicitly extract methodological patterns, unresolved contradictions, and areas of consensus.
	
	\item \textbf{Practical resources for researchers:} We compile a guide to the most commonly used public datasets (DAIC-WOZ, MODMA, RSDD) and standard evaluation metrics (accuracy, F1, AUC-ROC, MAE, RMSE, CCC), complete with their mathematical definitions and references to studies that employ them. This serves as a practical starting point for newcomers and a reference for experienced researchers.
	
	\item \textbf{Future roadmap and open challenges:} We highlight emerging trends—such as the dominance of GNNs in neuroimaging and the rise of LLMs for multimodal fusion—alongside open challenges including the need for causal inference, improved interpretability, privacy-preserving learning (e.g., federated learning), and cross-lingual/cross-cultural generalization.
\end{enumerate}

The remainder of this paper is organized as follows. Section \ref{sec:preliminaries} defines key clinical and technical concepts. Section \ref{sec:taxonomy} presents our hierarchical taxonomy. Section \ref{sec:related_works} provides a detailed, structured review of the literature. Sections \ref{sec:datasets} and \ref{sec:metrics} describe the prominent public datasets and evaluation metrics, respectively. Section \ref{sec:methodology} explains our paper selection and categorization process. Section \ref{sec:future_directions} discusses promising future research directions, and Section \ref{sec:conclusion} concludes the paper.

\section{Preliminaries}
\label{sec:preliminaries}
In this section, we provide essential definitions and background information relevant to our study, grounded in established clinical and technical literature.

\subsection{Depression}
Depression is a common and serious medical illness that negatively affects how you feel, the way you think, and how you act \cite{american2018what}. It is characterized by persistent feelings of sadness and a loss of interest or pleasure in previously rewarding or enjoyable activities. Beyond these core emotional symptoms, it can also disturb sleep and appetite, and lead to tiredness and poor concentration \cite{world2023depressive}. Unlike temporary mood fluctuations in response to life's challenges, the symptoms of clinical depression can be long-lasting and severe enough to significantly impair an individual's ability to function at work, at school, or within the family \cite{american2018what, world2023depressive}.

\subsection{Depressive Disorder}
Depressive disorder is a clinical term for a group of mood disorders where depression is the main feature. The two most common types are Major Depressive Disorder (MDD) and Persistent Depressive Disorder (dysthymia) \cite{american2022dsm5}.
{\em Major Depressive Disorder} involves discrete episodes of at least two weeks' duration involving clear-cut changes in affect, cognition, and neurovegetative functions that cause significant distress or impairment \cite{american2022dsm5}.
{\em Persistent Depressive Disorder}, or dysthymia, is a more chronic form of depression, characterized by a depressed mood that lasts for at least two years, though it may be less severe than an episode of major depression \cite{american2022dsm5}. The diagnosis and classification of these disorders are formally defined by criteria in diagnostic manuals such as the Diagnostic and Statistical Manual of Mental Disorders, Fifth Edition \cite{american2022dsm5}.

\subsection{Graph Neural Network}
A Graph Neural Network (GNN) is a class of deep learning models designed specifically to perform inference on data structured as graphs \cite{DBLP:conf/iclr/KipfW17,wu2020comprehensive,DBLP:journals/natmi/Chehreghani22,DBLP:journals/tjs/ZohrabiSC24,DBLP:journals/corr/abs-2401-01384,10.1145/3700790,DBLP:conf/iclr/KanatsoulisCJLR25,HOSEINNIA2025113615}. In a graph, information is stored in nodes (entities) and edges (relationships), and GNNs leverage this structure by passing and aggregating messages between neighboring nodes \cite{gori2005new}. This architecture allows them to effectively learn representations that capture both the features of the entities and the intricate topology of their connections. GNNs have proven highly effective in domains where relational data is key, such as social network analysis, molecular chemistry, and modeling brain functional connectivity networks \cite{wu2020comprehensive}.

\subsection{Large Language Model}
A Large Language Model (LLM) is a type of deep learning model, often based on the Transformer architecture, that is pre-trained on vast quantities of text data to understand, process, and generate human language \cite{zhao2023survey}. Seminal models like the original Transformer introduced the self-attention mechanism, enabling models to weigh the importance of different words in a sequence \cite{vaswani2017attention}, while subsequent models like BERT (Bidirectional Encoder Representations from Transformers) refined this by learning deep bidirectional representations from unlabeled text \cite{devlin2018bert}. LLMs are capable of performing a wide array of natural language processing (NLP) tasks, including text generation, question answering, and sentiment analysis, often with minimal fine-tuning.

\subsection{Electroencephalography Data}
Electroencephalography (EEG) is a non-invasive neurophysiological technique used to record the electrical activity generated by the brain via electrodes placed on the scalp \cite{niedermeyer2005electroencephalography}. The resulting EEG data captures the brain's spontaneous electrical potentials, which manifest as various rhythmic oscillations (brain waves) categorized by frequency bands such as delta, theta, alpha, and beta. These patterns reflect different states of consciousness and cognitive processes \cite{teplan2002fundamentals}. In clinical practice, EEG is a valuable tool for diagnosing and managing neurological conditions like epilepsy and sleep disorders, and it is increasingly used in psychiatric research to investigate biomarkers for conditions such as depression and schizophrenia \cite{al-farras2023review}.

\section{A Hierarchical Taxonomy for AI in Depression Detection}
\label{sec:taxonomy}

\begin{figure*}[!t]
	\centering
	\includegraphics[width=\linewidth]{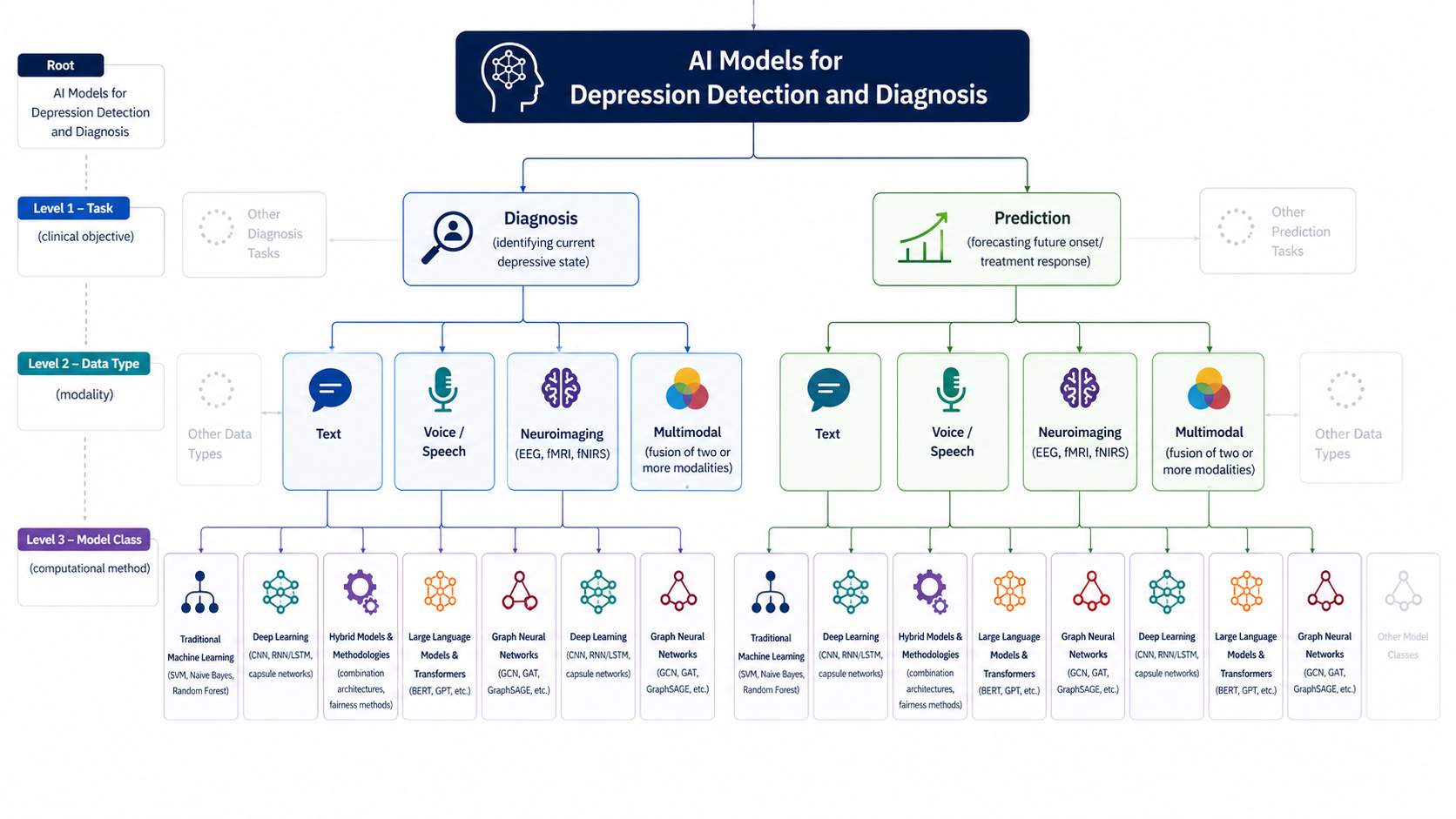}
	\caption{A hierarchical taxonomy of AI-based depression detection methods, organized across three levels: clinical task, data modality, and AI model.}
	\label{fig:taxonomy}
\end{figure*}

To provide a clear and structured overview of the vast body of literature in AI-based depression detection, we have developed a hierarchical taxonomy, illustrated in Figure~\ref{fig:taxonomy}. This framework organizes the 55 papers reviewed in this survey into three distinct levels: the {\em clinical task}, the {\em data type utilized}, and the {\em AI model employed}. This framework organizes the 55 papers reviewed in this survey into three distinct levels:
the {\em clinical task}, the {\em data type utilized}, and the {\em AI model employed}.
This section defines the categories within each level and presents tables that classify the reviewed literature accordingly, serving as a guide for the in-depth discussion that follows in Section \ref{sec:related_works}.

\subsection{Classification by Task}
The first level of our taxonomy distinguishes papers based on their primary clinical objective. We identify two fundamental tasks in the literature:

\begin{itemize}
	\item {\em Diagnosis:} This task involves the use of AI models to identify the current depressive state of an individual. This includes classifying subjects as depressed versus non-depressed or assessing the severity of their condition based on established clinical criteria.
	\item {\em Prediction:} This task focuses on forecasting future mental health outcomes. Models are designed to predict the future onset of depression, a user's risk level over time, or their potential response to a specific treatment.
\end{itemize}

Table~\ref{tab:task_classification} presents the distribution of the reviewed papers across these two tasks.

\begin{table*}[!t]
	\centering
	\caption{Categorization of reviewed papers based on the primary task addressed.}
	\label{tab:task_classification}
	\resizebox{\textwidth}{!}{
		\begin{tabular}{|l|p{12cm}|}
			\hline
			\textbf{Task Type} & \textbf{Papers} \\
			\hline
			Diagnosis & \cite{8389299}, \cite{huang2024depression}, \cite{10510456}, \cite{nusrat2024multiclassdepressiondetection}, \cite{patil2023depression}, \cite{liu2024depression}, \cite{10.1145/3569580}, \cite{ZHU2022105815}, \cite{LIU2024101081}, \cite{9844754}, \cite{chen-etal-2024-depression}, \cite{zhang-etal-2024-llms}, \cite{10680255}, \cite{nguyen2022improvinggeneralizabilitydepressiondetection}, \cite{10354236}, \cite{10517391}, \cite{10313367}, \cite{9911215}, \cite{s22041545}, \cite{ZHANG2023107478}, \cite{li2024detecting}, \cite{electronics12245040}, \cite{9906405}, \cite{10384682}, \cite{10.3389/fpsyt.2023.1125339}, \cite{Ma2025}, \cite{wang2024hybridgraphneuralnetwork}, \cite{SUN2024105675}, \cite{teng2025enhancingdepressiondetectionchainofthought}, \cite{zhang2025explainabledepressiondetectionclinical}, \cite{zhang2025speechtragreliabledepressiondetection}, \cite{10834580}, \cite{yang2025frequencyfeaturefusiongraph}, \cite{Kim2025Neurobiologically}, \cite{cha2024mogammultimodalobjectorientedgraph}, \cite{Yan2025DECEN}, \cite{Zhu2025MTNet}, \cite{Naseem2025}, \cite{Lim2025}, \cite{cheong2025ufairuncertaintybasedmultimodalmultitask}, \cite{kwok2025machinelearningfairnessdepression} \\
			\hline
			Prediction & \cite{xing2024adaptive}, \cite{kuo2023dynamicgraphrepresentationlearning}, \cite{10675718}, \cite{10356486}, \cite{10268256}, \cite{bidja2019depressiongnn}, \cite{qin2023readdiagnosechatexplainable}, \cite{shah2025depression}, \cite{10889035}, \cite{PerezToro2025Biases}, \cite{Jiao2025DeepGraph}, \cite{kumar2025transformer}, \cite{ibrahimov2025depressionxknowledgeinfusedresidual} \\
			\hline
		\end{tabular}
	}
\end{table*}

\subsection{Classification by Data Type}
The second level of our taxonomy categorizes studies based on the data modalities they utilize. The choice of data is a critical factor that shapes the methodological approach. We identify four main categories:

\begin{itemize}
	\item {\em Text:} This modality includes written language from diverse sources such as social media platforms (Twitter, Reddit), clinical interview transcripts, and online health forums.
	\item {\em Voice:} This modality encompasses acoustic and paralinguistic features of speech, such as pitch, tone, energy, and speech rate, captured from audio recordings.
	\item {\em Neuroimaging:} This category refers to data that captures brain structure or function. Common examples in the literature include electroencephalography (EEG), functional magnetic resonance imaging (fMRI), and functional near-infrared spectroscopy (fNIRS).
	\item {\em Multimodal:} This approach involves the synergistic integration of two or more data types (e.g., text and audio from an interview, or EEG and eye-tracking data) to create a more holistic and robust model.
\end{itemize}

Table~\ref{tab:data_classification} shows the classification of the reviewed papers based on data type.

\begin{table*}[!t]
	\centering
	\caption{Categorization of reviewed papers based on the data modality utilized.}
	\label{tab:data_classification}
	\resizebox{\textwidth}{!}{
		\begin{tabular}{|l|p{12cm}|}
			\hline
			\textbf{Data Type} & \textbf{Papers} \\
			\hline
			Text & \cite{8389299}, \cite{10510456}, \cite{nusrat2024multiclassdepressiondetection}, \cite{patil2023depression}, \cite{liu2024depression}, \cite{10.1145/3569580}, \cite{nguyen2022improvinggeneralizabilitydepressiondetection}, \cite{10354236}, \cite{10313367}, \cite{10356486}, \cite{qin2023readdiagnosechatexplainable}, \cite{teng2025enhancingdepressiondetectionchainofthought}, \cite{shah2025depression}, \cite{zhang2025explainabledepressiondetectionclinical}, \cite{Merzougui2025}, \cite{10889035}, \cite{PerezToro2025Biases}, \cite{Yan2025DECEN}, \cite{Naseem2025}, \cite{kumar2025transformer}, \cite{ibrahimov2025depressionxknowledgeinfusedresidual} \\
			\hline
			Voice & \cite{huang2024depression} \\
			\hline
			Neuroimaging & \cite{ZHU2022105815}, \cite{9844754}, \cite{10680255}, \cite{10517391}, \cite{10268256}, \cite{bidja2019depressiongnn}, \cite{ZHANG2023107478}, \cite{electronics12245040}, \cite{9906405}, \cite{10384682}, \cite{10.3389/fpsyt.2023.1125339}, \cite{Ma2025}, \cite{wang2024hybridgraphneuralnetwork}, \cite{10834580}, \cite{yang2025frequencyfeaturefusiongraph}, \cite{Jiao2025DeepGraph}, \cite{Kim2025Neurobiologically}, \cite{kwok2025machinelearningfairnessdepression} \\
			\hline
			Multimodal & \cite{xing2024adaptive}, \cite{kuo2023dynamicgraphrepresentationlearning}, \cite{LIU2024101081}, \cite{chen-etal-2024-depression}, \cite{zhang-etal-2024-llms}, \cite{10675718}, \cite{9911215}, \cite{s22041545}, \cite{li2024detecting}, \cite{SUN2024105675}, \cite{zhang2025speechtragreliabledepressiondetection}, \cite{cha2024mogammultimodalobjectorientedgraph}, \cite{Zhu2025MTNet}, \cite{Lim2025}, \cite{cheong2025ufairuncertaintybasedmultimodalmultitask} \\
			\hline
		\end{tabular}
	}
\end{table*}

\subsection{Classification by AI Model}
The final level of our taxonomy classifies papers based on the core AI model or methodology employed. This highlights the evolution of computational techniques in the field. We define five major categories:

\begin{itemize}
	\item {\em Traditional Machine Learning:} This category includes foundational models like Support Vector Machines (SVM), Naive Bayes, and Random Forests, which typically rely on manually engineered features from the data.
	\item {\em Deep Learning :} This refers to the application of a single, deep learning architecture, such as a Convolutional Neural Network (CNN), a Recurrent Neural Network (RNN) and etc.
	\item {\em Hybrid Models \& Methodologies:} This class includes studies that combine multiple distinct model architectures (e.g., CNN and LSTM) to leverage their complementary strengths. It also encompasses works focused on complex methodologies like algorithmic fairness and bias mitigation.
	\item {\em Large Language Models (LLMs) \& Transformers:} This category is dedicated to models based on the Transformer architecture, including seminal models like BERT and GPT, as well as their modern variants. These models are pre-trained on vast text corpora and are adept at understanding linguistic nuances.
	\item {\em Graph Neural Networks (GNNs):} This class consists of models specifically designed to operate on graph-structured data. In this context, they are primarily used to model brain connectivity networks or social interaction graphs.
\end{itemize}

Table~\ref{tab:model_classification} provides a detailed breakdown of the reviewed literature according to the AI models used.

\begin{table*}[!t]
	\centering
	\caption{Categorization of reviewed papers based on the primary model architecture employed.}
	\label{tab:model_classification}
	\resizebox{\textwidth}{!}{
		\begin{tabular}{|l|p{12cm}|}
			\hline
			\textbf{Model Type} & \textbf{Papers} \\
			\hline
			Traditional Machine Learning & \cite{8389299}, \cite{nusrat2024multiclassdepressiondetection}, \cite{nguyen2022improvinggeneralizabilitydepressiondetection}, \cite{10356486} \\
			\hline
			Hybrid Models \& Methodologies & \cite{10.1145/3569580}, \cite{Yan2025DECEN}, \cite{wang2024hybridgraphneuralnetwork}, \cite{kwok2025machinelearningfairnessdepression}, \cite{SUN2024105675}, \cite{cha2024mogammultimodalobjectorientedgraph}, \cite{Zhu2025MTNet}, \cite{Lim2025}, \cite{cheong2025ufairuncertaintybasedmultimodalmultitask}, \cite{ibrahimov2025depressionxknowledgeinfusedresidual} \\
			\hline
			Deep Learning & \cite{10510456}, \cite{patil2023depression}, \cite{liu2024depression}, \cite{9906405} \\
			\hline
			LLMs \& Transformers & \cite{huang2024depression}, \cite{chen-etal-2024-depression}, \cite{zhang-etal-2024-llms}, \cite{10675718}, \cite{10354236}, \cite{10313367}, \cite{li2024detecting}, \cite{qin2023readdiagnosechatexplainable}, \cite{teng2025enhancingdepressiondetectionchainofthought}, \cite{shah2025depression}, \cite{zhang2025explainabledepressiondetectionclinical}, \cite{Merzougui2025}, \cite{10889035}, \cite{PerezToro2025Biases}, \cite{zhang2025speechtragreliabledepressiondetection}, \cite{Naseem2025}, \cite{kumar2025transformer} \\
			\hline
			Graph Neural Networks (GNNs) & \cite{ZHU2022105815}, \cite{xing2024adaptive}, \cite{kuo2023dynamicgraphrepresentationlearning}, \cite{LIU2024101081}, \cite{9844754}, \cite{10680255}, \cite{10517391}, \cite{9911215}, \cite{10268256}, \cite{s22041545}, \cite{bidja2019depressiongnn}, \cite{ZHANG2023107478}, \cite{electronics12245040}, \cite{10384682}, \cite{10.3389/fpsyt.2023.1125339}, \cite{Ma2025}, \cite{10834580}, \cite{yang2025frequencyfeaturefusiongraph}, \cite{Jiao2025DeepGraph}, \cite{Kim2025Neurobiologically} \\
			\hline
		\end{tabular}
	}
\end{table*}

\section{Related Work}
\label{sec:related_works}

The landscape of mental health diagnosis has evolved significantly with the advent of computational methodologies that leverage various data modalities. As traditional psychiatric practices face challenges in accurately and promptly identifying depressive disorders, researchers are increasingly turning to innovative approaches that harness the power of technology. This section reviews the related works in the field of diagnosis and prediction of depression, emphasizing the integration of textual data, voice analysis, neuroimaging, and multi-modals. By examining the diverse computational techniques employed, we highlight how these methodologies not only enhance diagnostic accuracy but also facilitate earlier intervention. The subsections that follow delve into specific modalities and methods, from traditional Machine Learning approaches  to advanced deep learning architectures, illustrating the breadth of research aimed at improving mental health outcomes through systematic analysis and interpretation of data.

\subsection{Depression Diagnosis}
Diagnostic models focus on determining an individual’s current depressive state. These methods analyze multiple data modalities to classify subjects as depressed or non-depressed and to assess the severity of their condition according to established clinical criteria.

\subsubsection{Text-Based Diagnosis}
Text remains one of the most widely used data modalities, leveraging content from social media, clinical interviews, and online forums.

\paragraph{Traditional Machine Learning}

Early approaches in text-based depression detection rely on traditional machine learning models with engineered features. These foundational methods establish the viability of using linguistic patterns for mental health assessment.
One prominent study \cite{8389299} presents a systematic approach for detecting depression from Twitter feeds using natural language processing (NLP). Tweets are gathered via the Twitter API with a curated list of keywords linked to poor mental well-being, requiring app authentication with consumer and access tokens.
A comprehensive preprocessing pipeline—comprising link and non-ASCII removal, tokenization, stemming, stop-word elimination, and part-of-speech tagging—refines the data. Finally, the cleaned text is vectorized to train and test machine learning classifiers (support vector machines and Naive Bayes), with performance assessed by F1-score and accuracy.

In a related multi-class depression detection approach, researchers build a custom tweet dataset to predict five depression subtypes—Bipolar, Psychotic, Atypical, Postpartum, and Major Depressive Disorder. They create psychiatrist-verified lexicons of indicative phrases and use Apify to scrape matching tweets. A rigorous manual annotation process then filters the corpus to include only tweets that explicitly convey personal experiences of depression, resulting in a high-quality, contextually grounded dataset for further analysis \cite{nusrat2024multiclassdepressiondetection}.

Further advancing the generalizability of these models, another work \cite{nguyen2022improvinggeneralizabilitydepressiondetection} develops a depression detection model grounded in the PHQ-9 clinical questionnaire. The methodology involves developing a depression detection model grounded in the PHQ9 clinical questionnaire, which is commonly used for depression screening. The approach consists of two main components: a questionnaire model that detects symptoms from PHQ9 and a depression detection model that predicts depression based on symptom presence in social media posts. The models range from rule-based pattern matching to a BERT-based classifier, progressively relaxing constraints to allow more flexibility in learning. The dataset consists of three Reddit-based depression detection datasets: RSDD, eRisk2018, and TRT. These datasets differ in construction methodologies, such as self-reported diagnoses or participation in mental health forums. A weakly supervised approach is used to label symptoms by leveraging regular expressions, sentiment models, and heuristics, ensuring the model generalizes well across datasets while remaining interpretable.

\paragraph{Hybrid Models \& Methodologies}
Several studies combine different deep learning architectures to leverage their complementary strengths for text-based diagnosis. One such work \cite{10.1145/3569580} presents a method for detecting depression through the analysis of social media text using a hybrid deep learning model known as Fasttext Convolution Neural Network with Long Short-Term Memory (FCL). This approach aims to improve early detection of depression, which is critical for timely intervention. The FCL model leverages Fasttext embeddings to enhance word representation, addressing limitations of traditional methods by capturing semantic information and out-of-vocabulary words. It combines Convolutional Neural Networks (CNN) for global feature extraction and Long Short-Term Memory (LSTM) networks for understanding local dependencies in text. The methodology involves data cleaning and preprocessing of a dataset containing tweets/posts labeled for depression, followed by the application of the FCL model, which includes padding strategies to ensure consistent input sizes. The model’s performance is evaluated against existing state-of-the-art methods, demonstrating higher accuracy in detecting depression from social media content.

In \cite{Yan2025DECEN}, the Depressive Emotion–Context Enhanced Network (DECEN) detects depression from about 40 000 Sina Weibo posts—including a DSM-5–annotated subset of term-level depressive emotions. DECEN comprises: (i) a BiLSTM+CRF Depressive Emotion Recognition module for semantic–syntactic extraction; (ii) a BERT+attention Emotion-Context Enhanced Representation module; and (iii) a BiLSTM classification layer. By modeling specific depressive emotions (e.g., anhedonia, suicidal ideation) and their contextual relations, DECEN outperforms generic sentiment or embedding-based models in accuracy, precision, and robustness to sarcasm and variable text lengths.  

\paragraph{Deep Learning}

With the advent of deep learning, researchers begin to utilize architectures like Recurrent Neural Networks (RNNs) and Convolutional Neural Networks (CNNs) to automatically learn feature representations from text.
A notable study \cite{10510456} investigates the identification and intensity assessment of depression through tweets on Twitter, employing artificial intelligence and deep learning techniques, specifically long short-term memory (LSTM) models. The dataset comprises 95,322 tweets labeled as “non-depressed” or “depressed,” with the latter further classified into “mild,” “moderate,” and “severe” categories based on clinical criteria from the DSM-5. Data collection focuses on tweets containing specific hashtags associated with depressive symptoms while excluding uplifting or irrelevant content. Emotional and semantic scores are calculated using the Valence Aware Dictionary and Sentiment Reasoner (VADER) and latent semantic indexing (LSI) methods, respectively. The final depression intensity score is normalized and categorized into four classes.

In \cite{patil2023depression}, Patil et al. develop a system to detect and assess depression levels from medical data by analyzing unstructured text with a convolutional neural network (CNN). To address the lack of a large, publicly available benchmark for depression analysis, the pipeline begins with text preprocessing and proceeds to feature extraction and classification. The model treats text as a one-dimensional matrix and employs one-dimensional convolutional layers to capture local patterns. A word embedding layer first transforms tokens into dense vectors, which the convolutional network then processes to extract salient features. Pooling layers reduce dimensionality while retaining key information, and fully connected layers integrate these features into a unified representation. Finally, a SoftMax classifier predicts depression levels based on the learned features \cite{patil2023depression}.

In \cite{liu2024depression}, Liu et al. propose DeCapsNet, which combines capsule networks with contrastive learning for interpretable, high-performance depression detection from online posts. The model overcomes prior limitations by extracting symptom capsules based on PHQ-9 descriptions, enabling hierarchical reasoning. DeCapsNet first assigns each post a risk score—computed by its similarity to PHQ-9 symptoms—to select the most relevant samples. Capsule layers then generate symptom and depression capsules, aggregating symptom features into a higher-level representation for classification. A contrastive learning objective further refines embeddings by pulling similar instances together and pushing different classes apart. Evaluated on eRisk2018, RSDD, and TRT datasets, DeCapsNet outperforms baseline models in both within- and cross-dataset settings. Its loss function combines dynamic routing and contrastive learning losses to enhance overall performance \cite{liu2024depression}.

\paragraph{LLMs \& Transformers}
The current state-of-the-art in depression detection is dominated by Large Language Models (LLMs) and Transformer architectures, which leverage pre-trained knowledge to capture the nuanced, contextual nature of language.
Chen et al. \cite{chen-etal-2024-depression} propose a Structural Element Graph (SEGA) for depression detection in clinical interviews. SEGA employs a directed acyclic graph to model interactions among interview components—questions, transcripts, audio, and video—structured according to expert knowledge to ensure meaningful feature extraction while reducing noise.
To address data scarcity, the authors introduce an LLM-guided data augmentation strategy that generates synthetic responses and rephrased transcripts. Contrastive learning is then applied to refine representations by drawing similar instances closer and separating dissimilar ones.
They evaluate SEGA on two real-world multimodal clinical interview corpora: DAIC-WOZ (English interviews labeled with PHQ-8) and EATD (Chinese interviews labeled with SDS). Although both datasets are small due to privacy constraints, augmenting them with LLM-generated responses enhances training robustness and improves depression detection performance \cite{chen-etal-2024-depression}.

Similarly, Zhang et al.\ \cite{zhang-etal-2024-llms} integrate speech signals into Large Language Models using Acoustic Landmarks. Their method proceeds in three stages: (1) extracting phonetic landmarks from raw speech, (2) fine-tuning the LLM with cross-modal instructions to interpret these landmarks, and (3) applying P-Tuning for depression classification. They evaluate on the DAIC-WOZ corpus of clinical interviews and mitigate data scarcity and class imbalance through sub-dialogue shuffling. By combining text inputs with acoustic landmarks, the approach achieves a state-of-the-art F1 score of 0.84, demonstrating that lightweight landmark features can rival more complex deep-learning–based speech representations.
Other work applies transformer networks—such as BERT, GPT-3.5, and ChatGPT-4—to clinical interviews by augmenting traditional datasets with simulated data to preserve privacy and boost performance \cite{10354236}. Using DAIC-WOZ and Extended-DAIC, these studies focus on text-based modalities, preprocess inputs to address class imbalance

In \cite{10313367}, the authors propose a language-model-only framework for automated depression assessment using the Extended DAIC-WOZ (E-DAIC) corpus, which features semi-clinical interviews.
The pipeline begins with data preprocessing, where OpenAI's Whisper model transcribes incomplete or noisy audio into accurate text.
Feature extraction consists of two stages: first, GPT-3.5-Turbo transforms the interview transcripts to highlight depression-relevant patterns; second, DepRoBERTa—a RoBERTa model fine-tuned on PHQ-labeled data—classifies depression severity.
Transcripts annotated with PHQ scores are then used to train a Support Vector Regression (SVR) model, with hyperparameters optimized via GridSearchCV.
This methodology demonstrates how advanced NLP techniques can improve the accuracy of automated depression detection and establishes a foundation for future research in the field \cite{10313367}.

Other research \cite{10354236} applies transformer-based networks—including BERT, GPT-3.5, and ChatGPT-4—to clinical interview transcripts. To address privacy concerns and augment limited corpora, the authors generate simulated data, improving depression recognition from linguistic features. While the framework can handle multimodal inputs, ethical constraints restrict evaluation to text, leveraging PHQ-8 scores for ground-truth labeling. The study details a comprehensive preprocessing pipeline—covering transcript cleaning, tokenization, and class-balancing—and achieves state-of-the-art performance in automated depression detection. These findings underscore AI's transformative potential for early mental health intervention and pave the way for future diagnostics \cite{10354236}.

Zhang et al.\ \cite{zhang2025explainabledepressiondetectionclinical} propose RED (Retrieval-augmented generation for Explainable Depression detection), a transparent method for depression detection from clinical interview transcripts. RED relies solely on text modality and grounds its predictions in retrieved evidence to prevent hallucinations.
The model is evaluated on DAIC-WoZ, a structured corpus of interview transcripts labeled with PHQ-8 scores. It employs an adaptive Retrieval-Augmented Generation framework, first generating personalized queries via an LLM to infer each participant’s profile and then retrieving context-sensitive dialogue segments.
Retrieved evidence is enriched by a social intelligence module that integrates relevant knowledge from the COKE cognitive knowledge base using an event-centric retriever. This augmented evidence supports both the depression prediction and the generation of interpretable explanations.
By tailoring retrieval to individual contexts and incorporating psychologically relevant knowledge, RED outperforms neural and post-hoc LLM baselines in accuracy and explanation quality. These results highlight RED’s potential for trustworthy, personalized mental health assessments \cite{zhang2025explainabledepressiondetectionclinical}.

In \cite{10313367}, the authors introduce a language-model-only framework for automated depression detection using the Extended DAIC-WOZ (E-DAIC) corpus. E-DAIC comprises semi-clinical interviews conducted by the virtual agent Ellie in a Wizard-of-Oz setting, involving 275 participants (170 males, 105 females) split into 163 training, 56 development, and 56 test instances. Each interview, lasting approximately 20 minutes, is labeled with a PHQ-8 score to assess depression severity.
The pipeline begins with data preprocessing: Whisper transcriptions of incomplete or noisy audio are corrected using OpenAI's Whisper model. Feature extraction employs a two-stage strategy to prevent overfitting. First, GPT-3.5-Turbo transforms transcripts by generating prompts that highlight depression-relevant features. Second, DepRoBERTa—a RoBERTa variant fine-tuned on these enhanced transcripts—classifies depression severity into “none”, “moderate”, or “severe”.
Finally, the PHQ-8 annotations guide the training of a Support Vector Regression (SVR) model to predict continuous PHQ scores, with hyperparameters optimized via GridSearchCV.  

In \cite{teng2025enhancingdepressiondetectionchainofthought}, Teng et al. introduce a novel Chain-of-Thought (CoT) prompting strategy for Large Language Models (LLMs) to improve interpretability and clinical alignment in text-based depression detection. The approach uses the E-DAIC dataset, which comprises PHQ-8–annotated transcribed clinical interviews.
The detection task is decomposed into four structured reasoning stages: 
\begin{enumerate}
	\item Emotion analysis to identify affective states.  
	\item Binary classification to detect the presence of depression.  
	\item Causal reasoning to uncover contributing and protective factors.  
	\item Severity assessment aligned with PHQ-8 scoring.
\end{enumerate} 
This explicit reasoning framework mirrors clinical diagnostic workflows and captures nuanced symptoms, such as anhedonia expressed via negated positives. CoT prompts are applied to multiple LLMs—including GPT-4o, Qwen2.5, and DeepSeek R1—and yield significant improvements in concordance correlation coefficient (CCC) and mean absolute error (MAE). Notably, GPT-4o with CoT achieves CCC = 0.732 and MAE = 3.37. Ablation studies confirm that explicit CoT prompting enhances even inherently reasoning-capable models, underscoring its value for reliable, interpretable mental health assessments \cite{teng2025enhancingdepressiondetectionchainofthought}.

\textbf{\textit{Synthesis.}} Across text-based diagnosis, a clear evolutionary pattern emerges: from traditional machine learning with handcrafted features to deep learning (LSTM, CNN, capsule networks) and now to LLMs and Transformers. A key methodological consensus is the increasing use of clinical grounding—particularly PHQ-9 symptom alignment—to improve both accuracy and interpretability. Hybrid architectures (e.g., FCL, DECEN) that combine local and global feature extractors consistently outperform single-model approaches. Contradictions remain: some studies achieve state-of-the-art with relatively small, fine-tuned BERT-style models, while others require massive LLMs (GPT-4) with complex prompting strategies; the trade-off between performance and computational cost is rarely discussed. Another unresolved tension is the generalizability across social media platforms and clinical interviews—models trained on Reddit often underperform on Twitter or clinical corpora. Consensus exists on two points: (1) class imbalance and dataset shift are major obstacles, and (2) interpretability (via attention maps, retrieval, or symptom capsules) is no longer optional but essential for clinical trust.

\subsubsection{Voice-Based Diagnosis}
Voice data contain rich paralinguistic and acoustic cues that strongly indicate depressive states.

\paragraph{LLMs \& Transformers}
Voice-based analysis has greatly benefited from pre-trained models capable of extracting high-quality features from raw audio. A recent paper introduces a novel artificial-intelligence approach for the early detection of depression using voice data, addressing the challenge of limited dataset size. The methodology employs the wav2vec 2.0 pre-training model as a feature extractor to derive high-quality voice representations from raw audio, specifically utilizing the DAIC-WOZ dataset, which comprises 189 dialogue recordings between patients and a virtual agent. The recordings are preprocessed to enhance data quality through voice segmentation, ensuring that only patient speech segments are analyzed. The audio files are randomly divided into training, validation, and test sets in a 6:2:2 ratio. The wav2vec 2.0 model, renowned for its effectiveness in speech processing, is fine-tuned to classify depression, achieving accuracies of 0.9649 for binary classification and 0.9481 for multi-class classification. The model architecture includes a feature encoder for capturing local acoustic patterns and a transformer module for modeling global context, with additional dropout layers to mitigate overfitting during training. Overall, the proposed method demonstrates strong generalization capabilities and practical applicability for assisting healthcare professionals in early depression screening \cite{huang2024depression}.

\textbf{\textit{Synthesis.}} The literature on voice-based diagnosis is surprisingly sparse—only one study meets our inclusion criteria. That study demonstrates that a pre-trained Transformer for speech (wav2vec 2.0) can achieve excellent accuracy on the DAIC-WOZ dataset. The pattern is clear: transfer learning from large-scale speech recognition models is highly effective, even with modest amounts of clinical data. No contradictions can be identified due to the lack of competing studies. The consensus, though based on thin evidence, is that voice is a promising modality that deserves far more attention. Key open questions include the robustness of acoustic features across different recording conditions, the added value of voice over text alone, and the need for larger, publicly available voice depression datasets.

\subsubsection{Neuroimaging-Based Diagnosis}
Neuroimaging modalities, such as EEG and fMRI, offer direct insights into the neural correlates of depression, while graph neural networks (GNNs) excel at analyzing brain connectivity patterns.

\paragraph{GNNs}
GNNs have emerged as the dominant architecture for modeling the non-Euclidean structure of brain networks. One pioneering study introduced a more accurate and objective method for detecting depression, moving beyond traditional clinician–patient assessments. This approach uses resting-state electroencephalography (EEG) recordings from 27 patients with depression and 28 healthy controls to construct a brain functional network. Functional connectivity matrices are generated using Pearson correlation, from which four linear EEG features—activity, mobility, complexity, and power spectral density—are extracted. The authors propose the Graph Input Layer Attention Convolutional Network (GICN), which integrates a trainable weight matrix in the input layer and adopts the brain functional network as its adjacency matrix. In 10-fold cross-validation, GICN achieves a recognition accuracy of 96.50\%, outperforming existing methods. The model also identifies critical brain regions—specifically the temporal and parietal–occipital lobes—that significantly contribute to depression classification. Their methodology includes rigorous EEG acquisition and preprocessing to ensure high-quality data, and the use of graph convolutional layers to enhance feature extraction and classification \cite{ZHU2022105815}.

Another study presents a novel depression recognition method using a graph neural network that integrates spatio‐temporal features extracted from functional near‐infrared spectroscopy (fNIRS) data. The authors collected fNIRS recordings from 96 participants and derived six statistical metrics—mean, maximum, minimum, variance, skewness, and kurtosis—as temporal node features, alongside channel connectivity measures (coherence and correlation) as spatial edge weights. Each subject’s data is modeled as a graph where nodes encode temporal attributes and edges represent spatial relationships, enabling the GNN to learn jointly from both feature types. Experimental results demonstrate that this approach surpasses traditional machine learning methods in accuracy, F1 score, and precision, achieving over a 10\% improvement in F1 score. The dataset, acquired at Renmin Hospital of Wuhan University, comprises 53‐channel fNIRS signals sampled at 100\,Hz during a verbal fluency task. This study underscores the efficacy of combining temporal and spatial information for automatic depression recognition \cite{9844754}.

In \cite{10680255}, the BrainIB framework is introduced—a graph neural network that leverages the information bottleneck (IB) principle to improve psychiatric disorder diagnosis from fMRI-derived connectivity data. Traditional classifiers often overfit and lack explainability, limiting their clinical utility. BrainIB overcomes these challenges by sampling informative subgraphs from whole-brain functional connectivity graphs, thereby enhancing generalization to unseen data.
Evaluated on three psychiatric cohorts—ABIDE, REST-meta-MDD, and SRPBS—BrainIB outperforms seven state-of-the-art methods in diagnostic accuracy. fMRI data are preprocessed with slice-timing correction, motion correction, and normalization, followed by extraction of functional connectivity (FC) matrices. The BrainIB architecture comprises:
\begin{enumerate}
	\item A subgraph generator that samples informative subgraphs from FC graphs.  
	\item A graph encoder that learns embeddings for each subgraph.  
	\item A mutual information estimator that quantifies the dependence between subgraphs and the original graphs. 
\end{enumerate}
In addition to its superior accuracy, BrainIB identifies subgraph biomarkers consistent with clinical findings, underscoring its potential for practical psychiatric applications \cite{10680255}.

Other research has focused on fusing static and dynamic brain networks. The Dynamic-Static Fusion Graph Neural Network (DSFGNN) is designed to diagnose Major Depressive Disorder (MDD) by integrating static and dynamic functional connectivity networks derived from resting-state fMRI data. It addresses challenges in brain connectivity modeling, representation learning, and interpretability. Specifically, static functional connectivity graphs are constructed using Pearson correlation coefficients computed over the entire fMRI time series, while dynamic graphs are generated via a sliding-window approach to capture temporal fluctuations. 
The methodology employs separate Graph Isomorphism Network encoders for the static and dynamic graphs to learn node-level representations, a spatiotemporal attention mechanism to aggregate these representations into a global graph embedding, and a temporal attention module with Gated Recurrent Units (GRU) to model temporal evolution. Additionally, DSFGNN incorporates causal disentanglement to isolate causal factors from non-causal ones—enhancing interpretability—and applies orthogonal regularization to promote diverse representations for improved generalization and stability.
The framework is evaluated on the Rest-meta-MDD dataset, a multi-site resting-state fMRI collection originally comprising 1,300 MDD patients and 1,128 healthy controls across 25 sites. After quality control and site selection, 832 MDD patients and 779 controls from 17 sites remain. All participants provided informed consent, and the study received ethical approval. Preprocessing steps included slice-timing correction, head-motion correction, nuisance regression, spatial normalization, and band-pass filtering (0.01–0.1 Hz) \cite{10517391}.  

Further research has explored multi-view learning and advanced graph neural network architectures. The Multi-View Graph Neural Network (MV-GNN) integrates spatial and topological information from functional connectivity (FC) networks derived from resting-state fMRI. This study leverages the REST-meta-MDD dataset, which includes data from 1,300 Major Depressive Disorder (MDD) patients and 1,128 healthy controls (HC) collected across 25 research groups in 17 hospitals throughout China \cite{ZHANG2023107478}.
The methodology extracts complementary views from FC networks to elucidate MDD’s neural correlates. Spatial connectivity patterns altered in MDD are identified via T-tests and refined using LASSO for feature selection while preventing information leakage. Model interpretability is enhanced through SHAP analysis, which underscores the cerebellum’s significance. Additionally, the topology of FC networks is examined at both global and local scales, revealing structural differences between MDD and healthy controls.

In a similar vein, the DepressionGraph framework employs a two-channel graph neural network (GNN) alongside a transformer-based architecture to capture time-varying information from brain functional connectivity networks. The model leverages the REST-meta-MDD consortium’s publicly available resting-state fMRI dataset, which comprises 533 subjects across 17 Chinese hospitals \cite{electronics12245040}.
For each subject, the fMRI time series is divided into discrete time slices, and functional connectivity networks (FCNs) are constructed where nodes correspond to brain regions of interest (ROIs) and edges reflect correlation coefficients between node features. A gated recurrent unit (GRU) network first encodes node features to integrate temporal context. These encoded features are then processed by a two-channel GNN: one channel extracts fine-grained local connectivity patterns, while the other captures coarse-grained global structures. A transformer module subsequently models temporal dynamics across the sequence of FCNs. Final classification into MDD or control is achieved with a softmax layer.

Other studies have introduced graph autoencoders (GAEs) for brain disorder diagnosis. This framework leverages graph convolutional networks to perform inductive embedding of functional connectivity (FC) networks derived from resting-state fMRI data. By preserving the non-Euclidean structure of brain graphs, it departs from traditional convolutional neural network approaches.
First, individualized FC graphs are constructed using the Ledoit–Wolf shrinkage estimator, followed by extraction of node-level features. A multi-layer GCN encoder then aggregates local neighborhood information via spectral convolution. In the unsupervised variant, a symmetric decoder reconstructs the original adjacency matrix from learned embeddings. In the supervised variant, the GCN encoder is trained end-to-end with a fully connected neural network (FCNN) that maps graph-level embeddings directly to diagnostic labels.
Because the GCN encoder is inductive, it generalizes across subjects’ networks without relying on a fixed population graph \cite{10384682}.

An ensemble graph neural network model has also been proposed for MDD diagnosis. This study leverages resting‐state fMRI data from the REST-meta-MDD collaboration, comprising 1,586 participants. Functional connectivity matrices are generated for each subject and represented as graphs, where nodes correspond to brain regions of interest (ROIs) and edges encode pairwise connectivity strengths.
To boost classification performance, the framework ensembles three base GNN architectures—Graph Convolutional Network, Graph Attention Network, and GraphSAGE—each processing the same graph input to learn complementary embeddings. The learned representations are concatenated and fed into a meta‐classifier with softmax activation to produce the final prediction. By combining diverse modeling biases, this ensemble approach improves generalization and accuracy over individual models, enabling discrimination between MDD and healthy controls as well as between first‐episode drug‐naïve (FEDN) and recurrent (REC) MDD subtypes \cite{10.3389/fpsyt.2023.1125339}.

The Adaptive Propagation Operator Graph Convolutional Network (APO-GCN) introduces an adaptive propagation operator to mitigate over-smoothing and better capture discriminative patterns in brain functional graphs. This framework is applied to resting-state fMRI data from the multi-site REST-meta-MDD Consortium. After constructing individual functional connectivity matrices from preprocessed BOLD signals, APO-GCN employs Chebyshev polynomial–based convolutions for computational efficiency and dynamically adjusts its propagation operator during training. This adaptive modulation of information flow between graph nodes preserves critical signals that conventional GCNs tend to homogenize. 
Model performance is validated using both 10-fold cross-validation and leave-one-site-out schemes, achieving classification accuracies up to 91.8\%. Interpretability is further enhanced through GNNExplainer, which identifies the most influential brain regions driving MDD classification \cite{Ma2025}.

Further advancements have introduced DSGNN, a dual-branch self-supervised graph neural network (GNN) that leverages contrastive learning for depression diagnosis using resting-state fMRI data.
In this framework, brain functional connectivity networks are modeled as graphs, and a self-supervised architecture comprising two parallel GNN branches is proposed. These branches are trained to maximize agreement between differently augmented views of the same graph via a contrastive loss. Data augmentations—perturbing node features or graph structure—encourage the model to learn invariant and robust representations.
Each branch consists of a GCN-based encoder followed by a projection head, and the resulting embeddings are contrasted in the latent space. After self-supervised pretraining, the shared encoder is fine-tuned for the downstream diagnostic classification task using a multilayer perceptron \cite{10834580}.

Further research has introduced a Frequency Feature Fusion Graph Network leveraging functional near-infrared spectroscopy (fNIRS) data for depression diagnosis. A new dataset comprising 1,086 subjects was collected to support this study. The proposed model is built on a Temporal Graph Convolutional Network (TGCN) that integrates spatial and temporal brain features extracted from fNIRS signals.
Temporal dynamics are enriched via a Discrete Fourier Transform (DFT), enabling the identification of frequency-domain biomarkers. These frequency features are then fused with original spatial features through a three-stage, phase-specific GCN architecture that independently models the silent (resting), task, and post-task periods.
Key innovations include a Temporal Fusion Module (TFM) that merges raw and frequency-based representations, and a Frequency Point-Biserial Correlation Attention Module (FAM) that assigns attention weights to the most discriminative channels and frequency bands for improved diagnostic accuracy \cite{yang2025frequencyfeaturefusiongraph}.

Finally, one study applies graph neural networks to causal connectomes derived from resting-state fMRI (rs-fMRI) in 1,296 young adults \cite{Kim2025Neurobiologically}. The methodology develops and compares GNN-based classifiers using both traditional functional connectomes—built from Pearson and partial correlations—and three causal connectome methods: the TwoStep algorithm, Granger causality, and regression dynamic causal modeling (rDCM). These graphs are processed by advanced GNN architectures, including MSGNN, GIN, and GAT. The principal innovation is that causal connectivity not only outperforms functional connectivity in classification but also enhances neurobiological interpretability. Through GNNExplainer and spatial correlations with PET ligand maps, the study links model-identified key nodes to neurotransmitter systems such as serotonin (5-HT1B) and dopamine (extrastriatal D2) \cite{Kim2025Neurobiologically}.
Another contribution is the node-aware contrastive graph learning (NCGL) framework, which advances self-supervised GNNs by modeling individual functional connectivity networks as graphs \cite{Naseem2025}. NCGL’s core innovation is a contrastive pretraining strategy that incorporates node-level discrimination into graph-level contrastive learning. It employs dual GCN encoders to process two augmented views of the same graph and aligns their embeddings with a hybrid loss: a graph-level contrastive term plus a node-level consistency regularizer. The node-level loss penalizes discrepancies in corresponding node embeddings across augmentations, enforcing fine-grained invariance.

\paragraph{Hybrid Models \& Methodologies}
Some studies have combined diverse modeling approaches to capture complementary aspects of neuroimaging data. For example, the Hybrid Graph Neural Network (HybGNN) framework leverages EEG signals for depression detection using a dual-branch architecture comprising a Common Graph Neural Network (CGNN) and an Individualized Graph Neural Network (IGNN).
The CGNN employs a fixed graph topology to learn shared depression-related patterns, whereas the IGNN dynamically constructs subject-specific graphs to capture individual abnormalities. To enhance hierarchical feature learning, HybGNN integrates a Graph Pooling and Unpooling Module (GPUM) that adaptively aggregates EEG channels into brain regions and reinjects this structural information back into the network.
Evaluated on the publicly available MODMA and HUSM EEG datasets, HybGNN outperforms state-of-the-art models, achieving accuracies of 95.42\% on MODMA and 93.50\% on HUSM \cite{wang2024hybridgraphneuralnetwork}.

Another significant contribution is the systematic investigation of machine learning fairness in EEG-based depression detection. This study represents the first comprehensive analysis of algorithmic bias in depression classification using electroencephalography (EEG) data. Three benchmark EEG datasets—Mumtaz, MODMA, and Rest—are evaluated; they differ in gender distribution, sampling rates, and electrode configurations. The methodology involves training three deep learning models for depression classification: Deep-Asymmetry (CNN-based), GTSAN (GRU with attention), and 1DCNN-LSTM (convolutional-recurrent network). Five bias mitigation strategies are applied at various stages: pre-processing (Mixup augmentation, data massaging), in-processing (loss reweighting, regularization for equalized odds), and post-processing (Reject Option Classification). The innovation lies in the comprehensive fairness-aware framework, which employs metrics such as statistical parity, equal opportunity, equalized odds, and equal accuracy to quantify bias and examine its sensitivity to model architecture, dataset characteristics, and mitigation techniques. The findings reveal persistent algorithmic and dataset biases across all models and datasets, with pronounced disparities along gender lines \cite{kwok2025machinelearningfairnessdepression}.

\paragraph{Deep Learning }
The SGP-SL model \cite{9906405} introduces a three-stage framework for detecting Major Depressive Disorder (MDD) from EEG signals: graph construction, self-attention graph pooling, and prediction. 
First, an adjacency matrix is constructed to represent relationships between EEG electrodes, capturing both local and global connectivity patterns. Next, this graph is refined through multiple self-attention graph pooling modules, which preserve critical information while reducing the graph to a unified vector representation. 
Finally, the pooled representation is fed into a multi-layer perceptron (MLP) that simultaneously predicts soft class labels and Patient Health Questionnaire-9 (PHQ-9) scores. The loss function combines classification and regression objectives, using Kullback–Leibler divergence for the classification loss, mean absolute error for the regression loss, and an additional disagreement loss term to strengthen the correlation between tasks. 
The model is evaluated on the MODMA dataset, which comprises resting-state EEG recordings from 53 subjects (24 MDD patients and 29 healthy controls) acquired with 128 electrodes \cite{9906405}.

\textbf{\textit{Synthesis.}} Neuroimaging-based diagnosis is dominated by Graph Neural Networks, which naturally model brain connectivity as graphs. A strong pattern is the shift from static functional connectivity (Pearson correlation over whole time series) to dynamic or hybrid static–dynamic representations, with the latter generally improving accuracy. Another trend is the incorporation of explainability (GNNExplainer, SHAP, causal disentanglement) and fairness analysis. A major consensus is that multi-site datasets (especially REST-meta-MDD) are essential for evaluating generalization; single-site studies are viewed as preliminary. Contradictions exist regarding the optimal graph construction method: some advocate for simple Pearson correlation, others for partial correlation, shrinkage estimators, or even causal connectomes. The best-performing models often use ensemble or self-supervised learning, but the computational overhead is rarely discussed. A clear consensus is that temporal–parietal and frontal regions are consistently important biomarkers across studies, lending biological plausibility to GNN-based approaches.

\subsubsection{Multimodal Diagnosis}
Multimodal approaches combine data from diverse modalities—such as text, audio, visual, and physiological signals—to develop more comprehensive and robust diagnostic models.

\paragraph{GNNs}
Graph neural networks excel at fusing features from multiple modalities represented within graph structures. The Local-Global Multimodal Fusion Graph Neural Network (LGMF-GNN) \cite{LIU2024101081} integrates functional MRI, structural MRI, and electronic health records (EHRs) to improve the objectivity of Major Depressive Disorder (MDD) diagnosis.
Tested on diverse, multinational cohorts, the LGMF-GNN achieved a classification accuracy of 78.75\% and an AUROC of 80.64\%, effectively distinguishing MDD subtypes and uncovering distinct brain connectivity patterns associated with the disorder.
In the proposed local-global architecture, we begin by constructing region-of-interest (ROI) graphs for each subject. A learnable adjacency matrix is derived from the ROI BOLD time series, and node attributes correspond to the columns of the functional connectivity matrix obtained from resting-state fMRI. A local ROI GNN then applies graph convolution with an attention mechanism to aggregate ROI information, while a gated recurrent unit (GRU) encoder generates regional embeddings from the time series. These embeddings feed into a graph generator that outputs a subject-specific adjacency matrix. The GNN predictor uses attention over this learned graph and its node features to produce local embeddings and classification results.

To incorporate anatomical and demographic data, we designed a Global-Local Transformer (GLT) encoder and a Pairwise Association Encoder (PAE), which transform T1-weighted MRI features and demographic variables into one-dimensional feature vectors. In the global subject GNN, functional, anatomical, and demographic features serve as node attributes across three subject graphs.
At the population level, modality-specific GCN blocks generate representations unique to each modality, while modality-common GCN blocks distill shared information. A multimodal attention block then refines these representations into a unified embedding. Finally, a multilayer perceptron (MLP) classifier produces the global prediction \cite{LIU2024101081}.

The MS$^2$-GNN model \cite{9911215} enhances Major Depressive Disorder (MDD) detection by integrating EEG and audio signals from the MODMA dataset. It comprises several key components designed for effective multimodal feature extraction and fusion.
Initially, Long Short-Term Memory (LSTM) networks extract task-oriented features from both audio and EEG modalities, capturing MDD-related characteristics while reducing dimensionality. A shared network then identifies common representations across modalities, and modality-specific networks encode unique features for each. The embeddings are reconstructed to preserve semantic integrity, and an attention mechanism fuses them into a compact multimodal representation for classification.
The dataset consists of multimodal samples with audio and EEG inputs denoted \(X_a\) and \(X_e\), respectively, alongside ground-truth labels. A GNN refines these representations by propagating information over a dynamically constructed affinity matrix, computed from the absolute differences between node embeddings. This strategy enables the model to learn inter-modality relationships without requiring a predefined adjacency structure \cite{9911215}.

The KARE framework integrates physical activity data from smart home sensors and cyberspace activity (e.g., internet logs) to detect early signs of depression in older adults. This study proposes the KARE (Knowledge graph-based Activity pattern Recognition for Early detection of depression) framework. The task is formulated as an anomaly detection problem, identifying deviations in individuals’ activity patterns that may signal depression.
The methodology involves constructing a knowledge graph (KG) to integrate heterogeneous data sources, resolving semantic inconsistencies and enabling a unified representation. The Cyber–Physical View Representation (CPVR) module maps both physical and cyberspace activities into the KG, while the Personalized Activity Pattern Recognition (PAPR) module employs a Graph Attention Network (GAT) to learn normal behavioral patterns and detect anomalies.
The innovation lies in the cross-domain fusion of data sources through graph-based representation learning, enhancing the accuracy and timeliness of depression detection in smart home environments \cite{s22041545}.

\paragraph{LLMs \& Transformers}
LLMs can serve as a powerful backbone for fusing multimodal data.
The DSE-HGAT model \cite{li2024detecting} detects depression from clinical interview transcripts by constructing a heterogeneous graph. The task is framed as a dialogue extraction problem, where each transcript comprises a sequence of questions and responses labeled as “depressed” or “non-depressed.”
The model consists of three primary components:
\begin{enumerate}
	\item Context encoder layer: Two BiLSTM networks capture both local utterance-level and global dialogue-level information. Semantic features (word embeddings) and syntactic features (part-of-speech tags and named entity recognition embeddings) are integrated.  
	\item Heterogeneous graph layer: A graph is constructed with five node types—word, utterance, speaker, type, and state—and four edge types representing relationships within the interview. A graph attention mechanism aggregates information across these nodes to model contextual dependencies.  
	\item Output layer: Representations of word and state nodes are aggregated and fed into a classifier to predict depression labels. To address class imbalance, a focal loss function is employed during training.  
\end{enumerate}
The model is evaluated on the DAIC-WOZ dataset \cite{li2024detecting}.

SpeechT-RAG \cite{zhang2025speechtragreliabledepressiondetection} is a novel framework for depression detection that integrates acoustic and textual modalities, with a focus on speech-derived temporal features. This approach is evaluated on the DAIC-WOZ corpus and addresses shortcomings of text-only large language models (LLMs) and retrieval-augmented generation (RAG) systems, which often fail to capture critical nonverbal cues.
The methodology begins by extracting acoustic landmarks—such as glottal onsets and voiced frications—from the speech signal. Durations between successive landmarks are computed to form temporal bigrams, which are then summarized into statistical features (e.g., mean, variance). These timing-based features serve as retrieval keys in a novel RAG process that selects relevant examples without additional fine-tuning.
Retrieved examples are converted into structured prompts and fed to LLMs (LLaMA2, LLaMA3) for classification. A Gaussian Process Classifier provides calibrated confidence estimates. By leveraging speech timing patterns as retrieval keys, SpeechT-RAG achieves higher F1 scores and lower calibration errors compared to text-only RAG and traditional fine-tuned models \cite{zhang2025speechtragreliabledepressiondetection}.

\paragraph{Hybrid Models \& Methodologies}
Hybrid models are particularly powerful in multimodal settings, as they can combine specialized architectures for each data type.
A two-stage graph neural network methodology is proposed for depression detection using audio signals \cite{SUN2024105675}. The model is evaluated on three diverse datasets—DAIC-WOZ, MODMA, and D-Vlog—that include clinical interviews and real-world recordings.
First, low-level frame-wise audio features (MFCCs, log F0, and constant-Q transform coefficients) are extracted and passed through a gated recurrent unit (GRU) network to capture temporal dependencies.
Next, a two-stage GNN is employed:
\begin{enumerate}
	\item Intra-audio graph: each audio frame is treated as a node, and graph attention layers aggregate contextual frame-level information.  
	\item Inter-audio graph: each node represents a full audio embedding, with edges weighted by cosine similarity and refined via an emotion-aware attention mechanism.
\end{enumerate}
A pre-trained CompactSER model generates high-level sentiment features, which are integrated with the GNN embeddings using a self-attention fusion module. This hybrid, hierarchical graph-based architecture significantly outperforms traditional methods \cite{SUN2024105675}.

Similarly, the Multimodal Object-Oriented Graph Attention Model (MOGAM) is a deep learning framework for detecting depression in social media vlogs by integrating visual, textual, and structural modalities. The dataset, curated from YouTube, comprises 4,767 vlogs categorized into daily, high-risk depression, and clinically diagnosed depression groups.
For each vlog, YOLOv5 identifies objects in video frames to construct an object co-occurrence graph, whose adjacency matrix is processed by a graph neural network (Graph Convolutional Network, Graph Attention Network, or GraphSAGE). The resulting graph features are fused with visual embeddings from a ResNet backbone and textual embeddings from KoBERT. A cross-attention mechanism within a transformer architecture integrates these modalities into a unified representation.
MOGAM’s object-oriented approach avoids reliance on human-centric cues such as facial expressions or body poses, thereby improving generalizability \cite{cha2024mogammultimodalobjectorientedgraph}.

The Multimodal Transformer Network (MTNet)  \cite{Zhu2025MTNet} is designed for mild depression detection by integrating electroencephalography (EEG) and eye-tracking data. The dataset comprises recordings from 49 adolescents, including 21 with mild depression and 28 healthy controls.
EEG signals are preprocessed into frequency-specific segments, then passed through spatial and temporal convolutional layers before being encoded by a multi-head self-attention transformer. Eye-tracking features are extracted from fixation patterns and incorporated at various fusion stages—early, intermediate, and late—between the two modalities.
The key innovation of MTNet lies in its exploration of fusion timing and method, demonstrating that intermediate fusion achieves the highest classification accuracy of 91.79\% \cite{Zhu2025MTNet}.

The lightweight cross‐modality model \cite{Lim2025} combines audio and text data to detect depression across three multilingual datasets: DAIC-WOZ (English), EATD-Corpus (Chinese), and the Korean Depression Dataset.
Audio signals are converted into Mel spectrograms and processed by an MLP‐Mixer, while textual transcripts are encoded using an XLM-RoBERTa transformer encoder. The resulting embeddings are fused via a cross‐attention mechanism that enables dynamic, bidirectional interaction between the two modalities.
A key innovation is the model’s streamlined architecture, which significantly reduces parameter count without sacrificing accuracy. Furthermore, the approach demonstrates strong cross‐lingual generalizability across all three languages \cite{Lim2025}.

The U-Fair framework \cite{cheong2025ufairuncertaintybasedmultimodalmultitask} is an uncertainty-based multimodal multitask learning model that leverages audio, visual, and textual modalities from the DAIC-WOZ and E-DAIC datasets for fairer depression detection. Grounded in the clinical structure of the PHQ-8 questionnaire, U-Fair treats each of its eight symptom-specific items as a separate task within a multitask learning setup.
To mitigate demographic bias, U-Fair introduces a gender-aware loss reweighting strategy based on aleatoric uncertainty. This dynamic reweighting adjusts task priorities according to observed task difficulty and gender-specific symptom distributions. The framework’s principled integration of gender-based uncertainty into loss optimization aligns with clinical diagnostic logic and improves fairness in model predictions \cite{cheong2025ufairuncertaintybasedmultimodalmultitask}.

\textbf{\textit{Synthesis.}} Multimodal diagnosis is the fastest-growing area, reflecting the recognition that depression manifests across verbal, acoustic, and physiological channels. Several patterns emerge: (1) GNNs and Transformers are the dominant fusion backbones, often combined with modality-specific feature extractors (LSTMs for audio, CNNs for EEG). (2) Fusion strategies vary widely—early, intermediate, and late—with intermediate fusion (e.g., cross-attention) generally outperforming others, though no single method is universally best. (3) Fairness and uncertainty quantification are recent but welcome additions. A clear consensus is that multimodal consistently outperforms unimodal, though the improvement margin depends on how complementary the modalities are. Contradictions exist about the optimal number of modalities: some studies show diminishing returns beyond two or three. Another unresolved issue is the sensitivity to missing modalities in real-world deployment. The DAIC-WOZ dataset remains the de facto benchmark, but its small size (189 interviews) raises concerns about overfitting. Emerging work on cross-lingual and privacy-preserving multimodal fusion points toward future directions.

\subsection{Depression Prediction}
Prediction models are designed to forecast the onset of depression or estimate a user’s risk level over time. These models are crucial for enabling early intervention and preventative care.

As noted in the introduction, the literature on prediction is considerably smaller than that on diagnosis; we review the available studies below and highlight this gap.

\subsubsection{Text-Based Prediction}
Analyzing longitudinal text data can reveal patterns predictive of future depressive episodes.

\paragraph{Traditional Machine Learning}
A depression prediction model for Arabic social media posts is developed using a psychologist-annotated Twitter dataset of 1,058 tweets, evenly divided into “depressed” and “non-depressed” classes \cite{10356486}. The proposed framework comprises four phases:
i) Preprocessing: cleaning and normalizing raw tweets.
ii) Medical concept extraction: mapping texts to UMLS concepts via quickUMLS.
iii) Feature representation: weighting extracted concepts with Bag-of-Words (BOW) and TF-IDF to generate numerical vectors.
iv) Classification: training five machine learning algorithms—Random Forest (RF), Naive Bayes (NB), Logistic Regression (LR), Support Vector Machine (SVM), and Stochastic Gradient Descent (SGD).

\paragraph{Hybrid Models \& Methodologies}
DEPRESSIONX is an explainable deep learning model for detecting depression severity from social media text. We evaluate it on two benchmark Reddit datasets (D1 and D2), where posts are labeled into four ordered severity levels—minimal, mild, moderate, and severe—according to DSM criteria and the Beck Depression Inventory (BDI) standards \cite{ibrahimov2025depressionxknowledgeinfusedresidual}.
The DEPRESSIONX architecture comprises:
\begin{enumerate}
	\item Multi-level textual encodings (word-, sentence-, and post-level).
	\item Knowledge infusion via a structured depression knowledge graph built from Wikipedia using the REBEL relation extraction model.
	\item Residual multi-head attention to fuse textual vectors with knowledge graph embeddings.
	\item GIN-GAT layers for modeling structural relationships within the knowledge graph.
\end{enumerate}
Severity predictions are produced through ordinal regression to respect the inherent ordering of depression classes.
Explainability is achieved by:
i) Attention maps that highlight critical words and sentences, and
ii) Visualization of optimized knowledge subgraphs revealing the key conceptual relationships driving each prediction.

\paragraph{LLMs \& Transformers}
Large Language Models (LLMs) are increasingly applied to predictive tasks on social media. Qin et al. \cite{qin2023readdiagnosechatexplainable} introduce an interactive, explainable system for depression detection using the TMDD (Twitter) and WU3D (Weibo) datasets. Their methodology represents each user as a collection of posts containing both text and images, and trains a vanilla depression detection model, \(F^*\), on the full dataset to generate predicted depression probabilities that serve as answer heuristics. To enhance interpretability, the system employs a Chain‐of‐Thought (CoT) framework that articulates the reasoning behind each diagnosis in a structured question–answer format. Additionally, a tweet selector filters relevant posts, an image descriptor converts visual content into textual descriptions, and a prompt manager interfaces with LLMs to deliver diagnostic results and facilitate interactive dialogue \cite{qin2023readdiagnosechatexplainable}.

Another study fine-tunes GPT-3.5 Turbo 1106 and LLaMA2-7B on a large-scale dataset of 40,000 Twitter posts, which comprises three subsets: a depression-labeled set (D1), a non-depression set (D2), and a depression-candidate set (D3). The methodology involves carefully fine-tuning the pre-trained LLMs with optimized hyperparameters to distinguish depressive from non-depressive content by leveraging linguistic cues and emoji sentiment analysis. This domain-specific refinement enables real-time, scalable monitoring of mental health signals on social media. The fine-tuned GPT-3.5 Turbo achieves a detection accuracy of 96.4\%, while LLaMA2-7B reaches 87.1\% \cite{shah2025depression}.

In a different direction, Kumar et al. \cite{kumar2025transformer} develop the AST-D system (Abstractive Summarization Transformer for Depression) to automate the summarization of depression detection literature. This work addresses the ever‐growing volume of research in depression detection by proposing an AI-powered pipeline that converts full-text scientific articles into concise abstracts. The methodology leverages DepressiLex, a dataset of 40 peer-reviewed papers, and fine-tunes multiple pre-trained transformer architectures—T5-Base, PEGASUS-Large, BART-large-CNN, Longformer-Encoder-Decoder (LED), and ProphetNet—using the original abstracts as reference summaries. Through rigorous comparative evaluation and domain-specific optimization, Longformer-LED emerges as the most effective model for summarizing complex mental health literature \cite{kumar2025transformer}.

Other studies explore biases in multilingual depression classification using the RADAR-MDD dataset, which comprises English, Spanish, and Dutch samples.
This study applies large language models (LLMs) to classify depression severity from speech-derived text, with an emphasis on biases related to language, age, and gender. We transcribe free-response speech recordings with Whisper and feed the resulting transcripts into several pre-trained LLMs—flanT5, RoBERTa, BERT, GPT-2, and mBERT—each augmented with a multilayer perceptron for binary classification of high versus low symptom severity.
Models are evaluated on both balanced and unbalanced datasets stratified by language, gender, and age to assess performance disparities. Our key contribution is a systematic analysis of demographic biases in multilingual depression classification; we show that balancing for age has a more pronounced effect on model performance than balancing for gender, and that language differences significantly impact accuracy \cite{PerezToro2025Biases,Naseem2025}.

Another work \cite{Merzougui2025} investigates symptom-based depression severity estimation using only textual data from the DAIC-WOZ dataset, which comprises 189 structured clinical interviews.
Instead of framing depression detection as a binary classification or regression task, this research estimates the severity of each symptom individually. The methodology compares both encoder-based and decoder-based large language models  using in-context learning (ICL) strategies—including zero-shot, few-shot, and chain-of-thought (CoT) prompting—as well as parameter-efficient fine-tuning (PEFT) techniques such as LoRA.
Models evaluated include ModernBERT, Mistral-7B, LLaMA-3, DeepSeek-R1, and proprietary architectures like Gemini-2.0-Flash. The key contribution demonstrates that zero-shot ICL configurations can outperform fine-tuned models and that reasoning-tuned variants such as DeepSeek-R1-8B achieve higher symptom-level accuracy \cite{Merzougui2025}.

\textbf{\textit{Synthesis.}} Prediction from text is less mature than diagnosis, but a clear pattern is the shift from traditional ML with handcrafted features to LLM-based in-context learning and fine-tuning. A notable consensus is that incorporating structured medical knowledge (UMLS concepts, depression knowledge graphs) or ordinal regression improves severity prediction. Chain-of-Thought prompting is emerging as a powerful technique for both accuracy and explainability. Contradictions exist regarding whether fine-tuning or zero-shot prompting yields better generalization; the answer appears to depend on dataset size and domain shift. Another tension is between using social media data (large but noisy) versus clinical interview transcripts (smaller but cleaner). The field increasingly recognizes that predicting future depression requires longitudinal data, yet most studies remain cross-sectional or use retrospective labels. A clear consensus is that multilingual and cross-cultural bias is a serious concern, and simple demographic balancing is insufficient.

\subsubsection{Neuroimaging-Based Prediction}
Longitudinal neuroimaging data can be used to predict the trajectory of depression or treatment response.

\paragraph{GNNs}
GNNs are widely applied to predict outcomes from neuroimaging data.
Study \cite{10268256} involves data collection from 79 participants, including 25 Android and 54 iPhone users (ages 18–25) at the University of Connecticut. Data types collected comprised weekly QUIDS survey results, clinical diagnoses obtained through standardized medical evaluations, Fitbit activity and sleep records, raw GPS trajectories, and categorical GPS data. Exploratory analysis utilized heatmap visualizations to examine spatial movement patterns. Graph Neural Networks (GNNs) were then applied by constructing participant‐similarity graphs, where edges were defined using various distance metrics—Euclidean similarity for continuous features and Levenshtein edit distance for categorical GPS sequences. The efficacy of each metric was assessed via K-Nearest Neighbors and spectral clustering; final clustering outputs were compared against clinical labels using F1 scores to identify the optimal metric for graph construction \cite{10268256}.

Another important study \cite{Jiao2025DeepGraph} integrates functional magnetic resonance imaging (fMRI) and electroencephalography (EEG) data from the EMBARC clinical trial to predict treatment responses in patients with major depressive disorder (MDD). Pre-treatment resting-state fMRI and EEG scans were obtained from 130 patients randomized to sertraline and 135 randomized to placebo. Functional connectivity matrices for each modality were constructed and augmented via Common Orthogonal Basis Extraction (COBE). These matrices were then encoded by parallel graph neural networks with dynamic, learnable weight and scaling matrices to capture spatial dependencies and enable interpretability of region‐specific contributions. Encoded feature embeddings were fused through a modality correlation maximization strategy to produce joint representations, which fed into a multilayer perceptron (MLP) for treatment‐outcome prediction. The framework identified key predictive brain networks—such as the frontoparietal control network and limbic system—associated with differential antidepressant and placebo responses \cite{Jiao2025DeepGraph}.

\textbf{\textit{Synthesis.}} Neuroimaging-based prediction is a small but important niche. The dominant approach is constructing participant similarity graphs and applying GNNs to cluster or predict outcomes. A key pattern is that combining multiple neuroimaging modalities (fMRI + EEG) yields better prediction of treatment response than either alone. The only clear consensus is that prediction is substantially harder than diagnosis, requiring larger longitudinal cohorts (e.g., EMBARC). Contradictions are hard to identify due to the small number of studies, but an open question is whether graph construction should be based on functional connectivity, structural similarity, or behavioral metadata. Future work needs standardized benchmarks for prediction tasks.

\subsubsection{Multimodal Prediction}
Combining multiple data streams over time offers a powerful approach for early and accurate prediction.

\paragraph{GNNs}
MentalNet is designed for early depression detection by integrating user interaction data and textual content from social media within a Deep Graph Convolutional Neural Network (DGCNN). The architecture consists of:
i) ego-network feature extraction: user interactions (replies, mentions, quote-tweets) are encoded with an LSTM autoencoder to derive node features,
ii) heterogeneous graph construction: nodes represent users and posts, with edges capturing interaction types,
iii) graph convolution: a DGCNN processes the heterogeneous graph using doubly stochastic normalization to stabilize learning, and
iv) classification: the final node embeddings are fed into a classifier to predict depression risk.
By stabilizing graph learning and capturing rich interaction patterns, MentalNet achieves up to 19\% improvements in precision, recall, and F1 score over existing methods, demonstrating its effectiveness for social media–based depression detection \cite{xing2024adaptive, kuo2023dynamicgraphrepresentationlearning}.

\paragraph{LLMs \& Transformers}
Large language models can integrate diverse multimodal inputs for predictive tasks.
Study \cite{10675718} introduces a framework that enhances mental health assessment by combining ensemble machine learning with LLMs. Using a dataset of 41,000 mental health entries, we compared AdaBoost, Voting, Bagging, and Random Forest models, identifying Random Forest as the most effective.
The predictive workflow comprises:
i) processing a user query with the Random Forest model to predict potential mental health issues, and
ii) forwarding the prediction via an API to Google Gemini, which generates personalized insights based on the predicted condition \cite{10675718}.
Another study employs emotion prompts to guide LLMs in fusing heterogeneous signals for enhanced multimodal depression detection accuracy \cite{10889035}.

\textbf{\textit{Synthesis.}} Multimodal prediction is the least represented category, but the few studies show a pattern: graph-based fusion of social interactions (MentalNet) and ensemble+LLM pipelines for personalized insights. A clear consensus is that combining multiple data sources (interaction graphs, text, demographics) improves early detection compared to single-modality prediction. However, no clear consensus exists on the optimal fusion architecture. A significant challenge is the lack of public longitudinal multimodal datasets with ground-truth future outcomes. Future work should prioritize collecting such resources.

\section{Datasets}
\label{sec:datasets}

The advancement of AI-driven depression detection depends critically on high-quality, well-annotated datasets. These resources are indispensable for training, validating, and benchmarking new computational models. Datasets in this field are diverse, reflecting the multifaceted nature of depression and spanning several data modalities.
We categorize these resources into three broad types:
\begin{itemize}
	\item \textbf{Clinical and Interview-Based Datasets:} Structured, multimodal data collected in controlled clinical or research settings, often including audio, video, and questionnaire responses.
	\item \textbf{Physiological and Neuroimaging Datasets:} Objective biological markers captured via modalities such as electroencephalography (EEG) and functional magnetic resonance imaging (fMRI).
	\item \textbf{Social Media Datasets:} Large-scale, unstructured text and metadata harvested from online platforms, reflecting real-world user behaviors and language.
\end{itemize}
This section introduces three of the most influential public datasets—each representative of one of the categories above—that have been instrumental in advancing research in AI-driven depression detection.

\subsection{Distress Analysis Interview Corpus (DAIC-WOZ)}
The Distress Analysis Interview Corpus (DAIC), particularly its Wizard-of-Oz (WOZ) subset, is a widely adopted benchmark in multimodal depression detection \cite{gratch2014distress}. It comprises conversational interviews designed to facilitate the automatic identification of psychological distress indicators—such as depression, anxiety, and PTSD. Each session pairs a human participant with Ellie, a virtual interviewer avatar controlled by a concealed human operator in a Wizard-of-Oz setup. This approach fosters naturalistic, empathetic dialogue, enabling dynamic, responsive interactions that elicit rich verbal and nonverbal cues associated with mental health status \cite{gratch2014distress}.

The DAIC-WOZ corpus comprises 189 interview sessions, divided into training (107), development (35), and test (47) sets. Sessions range from 7 to 33 minutes in length. Each session provides synchronized high-fidelity facial video, audio recordings, and time-aligned text transcriptions of the dialogue.
Each participant’s data is annotated with a Patient Health Questionnaire (PHQ-8) score to quantify depression severity, with a score of 10 or higher indicating moderate depression. A binary depression label based on this threshold is also included. Additionally, the corpus offers pre-extracted feature sets—facial action units from video and acoustic features (e.g., pitch, intensity) from audio—enabling researchers to work without extensive signal-processing expertise.
Owing to its structured design, clinical grounding, and rich multimodal features, DAIC-WOZ is an invaluable benchmark for developing and comparing models that fuse verbal and nonverbal cues. A sample of the transcribed data is shown in Table \ref{tab:daicwoz}.

\begin{table}[!t]
	\centering
	\caption{An illustrative sample from the DAIC-WOZ dataset transcript \cite{gratch2014distress}.}
	\label{tab:daicwoz}
	\begin{tabular}{|p{0.15\columnwidth}|p{0.75\columnwidth}|}
		\hline
		\textbf{Speaker} & \textbf{Utterance} \\ \hline
		Ellie (Agent)    & So, tell me about yourself. \\ \hline
		Participant 300  & I'm from Los Angeles. I'm a student at U S C. I'm twenty one. I love to cook. uhm I don't know what else. \\ \hline
		Ellie (Agent)    & Tell me about something you did recently that you really enjoyed. \\ \hline
		Participant 300  & I had some friends over for dinner last week. I made uhm homemade pasta and some salads and it was just a really lovely evening. \\ \hline
	\end{tabular}
\end{table}

\subsection{Multi-modal Open Dataset for Mental-disorder Analysis (MODMA)}
The MODMA dataset was developed to address the need for open, high-quality physiological and behavioral data in mental health research, with a focus on Major Depressive Disorder (MDD) \cite{cao2020modma}. It comprises a comprehensive corpus that includes both behavioral recordings (audio and video) and direct neurophysiological signals, which are often absent from publicly available datasets. Data were collected from 53 participants—24 diagnosed with MDD and 29 healthy controls—recruited from a university and a psychiatric hospital. The experimental protocol included multiple sessions, featuring a resting-state period and a task-based period during which participants viewed emotionally evocative video clips designed to elicit varying affective responses.

The MODMA dataset comprises four primary modalities for each participant:
\begin{enumerate}
	\item High-density 128-channel electroencephalography (EEG) recorded with a Biosemi ActiveTwo system, providing detailed information on brain dynamics.
	\item Audio recordings from clinical interviews conducted using the Hamilton Depression Rating Scale (HAMD).
	\item Eye movement data captured with Tobii Pro Glasses 2.
	\item Facial expression videos recorded throughout each session.
\end{enumerate}
In addition to the MDD diagnosis, each record is annotated with scores from multiple clinical scales, including the Patient Health Questionnaire (PHQ-9), the Generalized Anxiety Disorder scale (GAD-7), and the Beck Depression Inventory (BDI-II). The inclusion of high-density EEG data makes MODMA an essential resource for researchers developing graph neural network models to analyze brain connectivity or hybrid models that fuse neurophysiological signals with behavioral cues \cite{cao2020modma}. An overview of the data collected per participant is shown in Table \ref{tab:modma}.

\begin{table}[!t]
	\centering
	\caption{Data modalities collected per participant in the MODMA dataset \cite{cao2020modma}.}
	\label{tab:modma}
	\begin{tabular}{|l|l|}
		\hline
		\textbf{Data Type} & \textbf{Specification} \\ \hline
		EEG & 128-channel Biosemi ActiveTwo system \\ \hline
		Audio Interview & Recordings from clinical interviews (HAMD) \\ \hline
		Eye Movement & Tobii Pro Glasses 2 eye tracker data \\ \hline
		Facial Video & High-resolution video of facial expressions \\ \hline
		Clinical Scores & PHQ-9, GAD-7, BDI-II, and clinical diagnosis \\ \hline
	\end{tabular}
\end{table}

\subsection{Reddit Self-reported Depression Diagnosis (RSDD) Dataset}
Social media–derived datasets are essential for studying depression in naturalistic, real-world settings. The Reddit Self-Reported Depression Diagnosis (RSDD) dataset exemplifies this approach by enabling text-based depression detection with more reliable labels than keyword-based collection methods \cite{schrading2021difficulties}. RSDD defines its “depressed” class strictly as users who explicitly self-report a formal depression diagnosis, thereby distinguishing clinical cases from transient expressions of sadness.

The RSDD dataset is constructed by identifying Reddit posts containing explicit indicators of a formal depression diagnosis (e.g., “I was diagnosed with depression,” “my doctor diagnosed me with depression”). For each such user, the entire posting history is retrieved to form the depressed cohort. A matched control group is then assembled by selecting users with similar activity levels—such as comparable post and comment counts—from a broad range of non–mental-health subreddits. The final corpus comprises the posting histories of 9,921 depressed users and 107,316 controls, providing a large-scale resource for training text-based models. Although this approach improves ecological validity and label quality over simple keyword searches, it is constrained by the inherent uncertainty of self-reported diagnoses. Nevertheless, RSDD remains invaluable for developing scalable models on extensive, unstructured, and longitudinal text data \cite{schrading2021difficulties}. A sample is shown in Table \ref{tab:rsdd}.

\begin{table}[!t]
	\centering
	\caption{An illustrative example of posts from the RSDD dataset \cite{schrading2021difficulties}.}
	\label{tab:rsdd}
	\begin{tabular}{|p{0.15\columnwidth}|p{0.75\columnwidth}|}
		\hline
		\textbf{User Label} & \textbf{Post Snippet (from post history)} \\ \hline
		Depressed & "I was officially diagnosed with severe depression today. I don't know how to feel. It's a relief to have a name for it but now it feels so... real." \\ \hline
		Control & "Just finished building my new PC. The cable management was tough but it's running Cyberpunk 2077 on ultra settings, so worth it!" \\ \hline
	\end{tabular}
\end{table}

\section{Evaluation Metrics}
\label{sec:metrics}
Rigorous assessment and comparison of AI-based depression detection models require standardized evaluation metrics. Metric selection depends on the model’s primary objective, which typically falls into one of two categories: classification or regression. Classification tasks assign discrete labels (e.g., depressed vs.\ non-depressed), while regression tasks predict continuous severity scores.
In the following, we introduce the key metrics used in the depression detection literature for both tasks, present their mathematical definitions, and cite examples of their application.

\subsection{Metrics for Classification Tasks}
Classification models are most commonly evaluated using metrics derived from the confusion matrix, which tabulates the number of correct and incorrect predictions for each class. The core components of the matrix are True Positives (TP), True Negatives (TN), False Positives (FP), and False Negatives (FN).

\paragraph{Accuracy}
Accuracy measures the proportion of correct predictions among all samples:
\[
\mathrm{Accuracy} = \frac{TP + TN}{TP + TN + FP + FN}\,.
\]
Although easy to interpret, accuracy can be misleading on imbalanced datasets, since a model may achieve high scores by always predicting the majority class. In Zhu et al.\ \cite{ZHU2022105815}, accuracy is employed as the primary metric for EEG-based depression recognition.

\paragraph{Precision, Recall, and F1-Score}
These metrics provide a more nuanced perspective on model performance, especially when the underlying class distribution is imbalanced.

\begin{itemize}
	\item \textbf{Precision}, or positive predictive value, is the proportion of true positives among all positive predictions:
	\[
	\text{Precision} = \frac{TP}{TP + FP}.
	\]
	It answers the question, “Of all individuals flagged as depressed by the model, how many were actually depressed?” High precision is vital in clinical settings to minimize false alarms. This metric is used in the fNIRS-based depression detection study \cite{9844754}.
	
	\item \textbf{Recall}, also known as sensitivity or true positive rate, is the proportion of actual positives correctly identified by the model:
	\[
	\mathrm{Recall} = \frac{TP}{TP + FN}.
	\]
	It answers the question, “Of all individuals who are truly depressed, how many did the model correctly identify?” High recall is vital in clinical contexts to ensure few cases are missed. This metric is also used in the fNIRS study \cite{9844754}.
	
	\item \textbf{F1-Score} is the harmonic mean of Precision and Recall, providing a single score that balances both metrics. It is particularly useful for evaluating models on imbalanced datasets where both minimizing false positives and false negatives is important. The F1-score is a key evaluation metric in the multimodal study by \cite{cha2024mogammultimodalobjectorientedgraph}.
	\[
	\text{F1-Score}
	= 2 \times \frac{\text{Precision}\times\text{Recall}}{\text{Precision} + \text{Recall}}
	= \frac{2 \times TP}{2 \times TP + FP + FN}.
	\]
\end{itemize}

\paragraph{Specificity}, also known as the true negative rate, is the proportion of actual negatives correctly identified by the model. It complements recall by confirming the model can effectively identify healthy individuals. This metric is used in the DepressionGraph framework \cite{electronics12245040}.
\[
\text{Specificity} = \frac{TN}{TN + FP}.
\]

\paragraph{Area Under the ROC Curve (AUC-ROC)} measures a model’s discriminative ability independent of any specific threshold. The ROC curve plots the true positive rate (recall) against the false positive rate (1 – specificity) across all classification thresholds. The AUC represents the probability that the model will rank a randomly chosen positive instance higher than a randomly chosen negative one. This metric is a primary evaluation criterion for the LGMF-GNN model \cite{LIU2024101081}.
\[
\mathrm{AUC\text{-}ROC}
= \int_{0}^{1} \mathrm{TPR}(t)\,\mathrm{d}\bigl(\mathrm{FPR}(t)\bigr).
\]

\subsection{Metrics for Regression Tasks}
For tasks that predict a continuous depression severity score, such as a PHQ-8 or BDI score, evaluation metrics measure the average error or agreement between the predicted ($y'$) and true ($y$) scores for a set of $n$ samples.

\paragraph{Mean Absolute Error (MAE)}
MAE measures the average absolute difference between the predicted and actual depression severity scores. It is easy to interpret since it represents the average prediction error in the original units of the score. MAE is used by \cite{teng2025enhancingdepressiondetectionchainofthought} to evaluate the performance of LLMs in predicting PHQ-8 scores.
\[
\mathrm{MAE} = \frac{1}{n} \sum_{i=1}^{n} \bigl|y_i - y_i'\bigr|
\]

\paragraph{Root Mean Square Error (RMSE)}
RMSE is the square root of the average of the squared differences between predicted and actual scores. By squaring the errors, it gives significantly more weight to larger errors, making it a useful metric when large deviations are particularly undesirable. RMSE is used to evaluate the SGP-SL model's ability to predict PHQ-9 scores from EEG data \cite{9906405}.
\[
\mathrm{RMSE} = \sqrt{\frac{1}{n} \sum_{i=1}^{n} \bigl(y_i - y_i'\bigr)^2}.
\]

\paragraph{Concordance Correlation Coefficient (CCC)} 
CCC measures the agreement between predicted and true scores by evaluating both their linear correlation and their deviation from the 45° line of perfect concordance. A CCC of 1 indicates perfect agreement, –1 perfect disagreement, and 0 no agreement. It is more robust than Pearson correlation for assessing prediction accuracy and is used as a key metric in the Chain-of-Thought LLM study \cite{teng2025enhancingdepressiondetectionchainofthought}.
\[
\mathrm{CCC}
= \frac{2\,\rho\,\sigma_{y}\,\sigma_{y'}}
{\sigma_{y}^{2} + \sigma_{y'}^{2} + (\mu_{y} - \mu_{y'})^{2}},
\]
where \(\rho\) is the Pearson correlation coefficient, and \(\mu\) and \(\sigma^{2}\) denote the means and variances of the respective variables.

\section{Future Research Directions and Open Challenges}
\label{sec:future_directions}

The body of work reviewed in this survey highlights significant progress in applying AI to depression detection. Analysis of the current literature reveals several prominent trends: the dominance of graph neural networks for neuroimaging; the rise of large language models for text and multimodal analysis; a strong push toward multimodal fusion; and an emerging focus on explainability and fairness. While these trends demonstrate the field’s maturation, they also underscore several open challenges and promising avenues for future research.

\subsection{From Multimodal Correlation to Causal Fusion}
A clear trend is the move toward sophisticated multimodal fusion, combining everything from EEG and audio signals \cite{9911215} to fMRI, structural MRI, and health records \cite{LIU2024101081}. Current models excel at identifying correlations between these data streams; however, the next critical step is to move from correlation to causation. Future research should prioritize developing causal inference models that untangle the complex interplay among modalities. For example, can changes in brain connectivity (from fMRI) be shown to causally influence specific vocal features (from audio) or linguistic patterns (from text)? Answering such questions would not only improve model robustness but also deepen insights into the neurobiological underpinnings of depression—a direction already explored in unimodal neuroimaging studies \cite{Kim2025Neurobiologically}. Furthermore, there is a need for lightweight fusion models that are computationally efficient enough for real-time deployment on personal devices such as smartphones, a crucial step toward continuous, real-world mental health monitoring \cite{Lim2025}.

\subsection{Enhancing Model Interpretability for Clinical Trust}
As models grow more sophisticated, there is an increasing emphasis on ensuring their explainability and interpretability in clinical settings. This focus directly counters the black-box nature of many deep learning systems. Recent innovative approaches include developing retrieval-augmented generation (RAG) frameworks that ground predictions in textual evidence from clinical transcripts \cite{zhang2025explainabledepressiondetectionclinical}, employing capsule networks to extract symptom-specific features aligned with the PHQ-9 \cite{liu2024depression}, and designing GNNs capable of identifying clinically relevant brain subgraphs \cite{10680255}.

Despite this progress, future work must move beyond post-hoc explanations toward models that are inherently interpretable by design. Research should focus on developing systems that generate clear, evidence‐based explanations in natural language, understandable to both clinicians and patients. For example, a model could not only predict high depression risk but also state that its reasoning is based on “reduced speech rate, increased use of first‐person pronouns, and anomalous connectivity in the frontoparietal control network.” Validating these explanations with mental health professionals will be crucial for building the trust necessary for clinical adoption.

\subsection{Shifting from Diagnosis to Longitudinal Prediction and Intervention}
The majority of reviewed studies focus on diagnosing current depression. While this remains crucial, the next frontier is forecasting depression onset, trajectory, and treatment response over time. Achieving this requires gathering and analyzing longitudinal data from diverse sources—such as wearable sensors and periodic self-assessments \cite{bidja2019depressiongnn}. Although a handful of studies have begun predicting treatment response using neuroimaging \cite{Jiao2025DeepGraph}, this domain remains largely unexplored. Ultimately, these predictive models should be integrated into personalized, just-in-time intervention systems that deliver timely, targeted support—whether suggesting coping strategies or prompting users to seek professional help at the earliest warning signs.

\subsection{Addressing Data Scarcity, Privacy, and Diversity}
The scarcity of large-scale, high-quality, and demographically diverse datasets remains a major bottleneck in the field, driven by the sensitive nature of mental health data and the high costs of clinical data collection. Future research must adopt privacy-preserving machine learning approaches. For instance, federated learning can enable model training on decentralized data from multiple hospitals or user devices without compromising patient confidentiality. Moreover, fairness studies have highlighted demographic imbalances in existing datasets \cite{kwok2025machinelearningfairnessdepression}. Developing more sophisticated, clinically validated data augmentation techniques—especially for underrepresented groups—will be critical for building robust, equitable models that generalize effectively to the broader population.

\subsection{Cross-lingual and Cross-cultural Generalization}
Much of the current research, particularly LLM‐based work, relies on English‐language datasets such as DAIC‐WOZ; however, studies on multilingual corpora show that performance can vary substantially across languages and cultural contexts \cite{PerezToro2025Biases}. Depressive symptom expression is not universal, and models trained on one cultural group may not generalize to others. Future research should prioritize culturally aware model development and validation. This necessitates curating diverse, multilingual datasets and designing flexible architectures—either by enabling efficient adaptation to new languages or by learning universal, language‐agnostic representations of depression.

\section{Methodology}
\label{sec:methodology}

To compile and structure the literature for this survey, we employed a systematic and multi-stage methodology. The process was designed to ensure a comprehensive, relevant, and high-quality selection of papers, which were managed and tracked using a Google Sheet.

\subsection{Search Strategy}
Our primary database for literature collection was Google Scholar. We conducted a series of targeted searches using combinations of keywords to identify pertinent studies at the intersection of artificial intelligence and mental health. The core search terms included "depression detection" and "MDD detection," which were paired with a range of technical keywords such as "machine learning," "AI," "GNN," "LLM," and "DL" (Deep Learning). This strategy yielded an initial pool of 61 papers that formed the basis for our subsequent screening and selection process.

\subsection{Inclusion and Quality Assessment}
Each of the 61 papers from the initial pool was subjected to a rigorous quality assessment to determine its suitability for inclusion in this survey. We developed a 10-point scoring system to evaluate each paper across three key criteria:
\begin{itemize}
	\item \textbf{Relevance:} How directly the paper addressed the task of depression detection using AI.
	\item \textbf{Venue Quality:} The reputation and impact of the journal or conference where the paper was published.
	\item \textbf{Recency:} The publication date, with a preference for more recent work to reflect the current state of the field.
\end{itemize}
A strict filtering protocol was applied: any paper that scored below 5 out of 10 on even one of these criteria was excluded from further consideration. This comprehensive review process ensured that our final selection was both current and of high academic standing, ultimately resulting in the 55 papers that form the core of this survey.

\subsection{Categorization}
A key contribution of this survey is its unique hierarchical classification framework, designed to provide a structured and intuitive overview of the research landscape. After a thorough analysis, we identified three primary variant factors among the selected papers and used them to build our taxonomy. Each paper was assigned to a single primary category at each level to ensure a clear and non-redundant classification.

The categorization was performed in the following hierarchical order:
\begin{itemize}
	\item \textbf{Level 1: Task Type.} The first and most fundamental division was based on the primary goal of the research. Papers were classified into two major groups:
	\begin{itemize}
		\item {\em Diagnosis:} Works focused on identifying the current depressive state of an individual.
		\item {\em Prediction:} Works focused on forecasting the future onset or risk of depression.
	\end{itemize}
	\item \textbf{Level 2: Data Type.} Within each task, papers were further classified based on the data modality used, which was the second most significant variant factor. This level includes categories such as Text, Voice, Neuroimaging, and Multimodal data.
	\item \textbf{Level 3: Methodology and Model.} Finally, at the deepest level of the hierarchy, papers were grouped by the specific AI model or methodology employed. As this was the most variant factor, it led to the creation of distinct classes such as Graph Neural Networks (GNNs), Large Language Models (LLMs), and Hybrid Models.
\end{itemize}
This hierarchical categorization strategy allows for a nuanced and detailed analysis of the field, enabling readers to navigate the literature based on their specific interests in clinical tasks, data sources, or computational techniques.

\section{Conclusion}
\label{sec:conclusion}
This survey has provided a comprehensive and systematically structured overview of the application of Artificial Intelligence in the detection of depressive disorders. By reviewing 55 recent important studies, we have mapped the current landscape, highlighting the significant progress made in leveraging computational models to analyze a diverse range of data modalities. Our unique hierarchical taxonomy, which categorizes research first by clinical task (Diagnosis vs. Prediction), then by data type, and finally by model class, offers a novel framework for understanding the field's key trends and identifying specific areas of innovation.

Our review confirms a clear trend towards the adoption of sophisticated deep learning architectures. Graph Neural Networks have become the standard for modeling the complex, structured data of neuroimaging, while Large Language Models and Transformers are revolutionizing the analysis of text and multimodal interview data. Furthermore, we identified a growing and critical focus on addressing the practical challenges of clinical implementation, with an increasing number of studies dedicated to enhancing model explainability and ensuring algorithmic fairness. Despite these advancements, significant challenges remain, including the need to move from correlational to causal models, address data scarcity through privacy-preserving techniques, and ensure models are culturally and linguistically generalizable.

By synthesizing the state of the art and outlining these open challenges, this paper serves as a valuable resource for researchers, clinicians, and engineers. It not only provides a detailed map of what has been accomplished but also offers a clear roadmap for future research directions. Continued interdisciplinary collaboration will be essential to overcoming the existing hurdles and fully realizing the potential of AI to create objective, accessible, and effective tools that can transform mental healthcare.

\section*{Ethics Statement}
This paper is a literature review and does not involve new data collection or human subjects research.  
However, ethical considerations—including algorithmic fairness, demographic bias, and data privacy—are discussed in the reviewed studies (see Sections~\ref{sec:related_works} and~\ref{sec:future_directions}).  
No ethical approval was required for this review work (see Section~\ref{sec:methodology} for the paper selection process).

\bibliographystyle{plain} 
\bibliography{references}           

\end{document}